\documentclass{article}
\usepackage[square,numbers]{natbib}
\usepackage{authblk}
\usepackage{charter}
\usepackage{fullpage}
\usepackage{bbm}

\usepackage{hyperref}       % hyperlinks
\usepackage{url}            % simple URL typesetting
\usepackage{booktabs}       % professional-quality tables
\usepackage{amsfonts}       % blackboard math symbols
\usepackage{nicefrac}       % compact symbols for 1/2, etc.
\usepackage{microtype}      % microtypography
\usepackage[table]{xcolor}         % colors
\usepackage{amsmath,amssymb,amsthm,amsfonts}
\usepackage{latexsym,bbm,graphicx,float,mathtools}
\usepackage[export]{adjustbox}
\usepackage{capt-of}% or \usepackage{caption}
\usepackage{booktabs}
\usepackage{graphicx}
\usepackage{parskip}
\usepackage[ruled,vlined]{algorithm2e}

\usepackage[utf8]{inputenc}

\usepackage[noabbrev]{cleveref}
\usepackage{caption}
\usepackage{subcaption}
\usepackage{transparent}
\usepackage[inline]{enumitem}
\usepackage{multirow}
\usepackage{multicol}
\usepackage{xspace}
\usepackage{fancyvrb}       % for Verbatim environment
\usepackage{fvextra}        % improved verbatim with line breaking
\usepackage{tcolorbox}
\tcbuselibrary{breakable,skins}

\usepackage{abstract}

\newtcolorbox{promptbox}[1]{
  breakable,
  enhanced,
  colback=blue!3!white,
  colframe=blue!60!black,
  fonttitle=\bfseries\large,
  title=#1,
  arc=3pt,
  boxrule=1pt,
  left=10pt,
  right=10pt,
  top=8pt,
  bottom=8pt,
  before skip=8pt,
  after skip=8pt,
  drop shadow={xshift=0.5mm, yshift=-0.5mm, fill=black!20!white},
  width=\textwidth,
  before=\centering
}

\usepackage[subtle, mathdisplays=tight, charwidths=normal, leading=normal]{savetrees}

\title{\textbf{TAG: Thinking with Action Unit Grounding for Facial Expression Recognition}}

\author{
  Haobo Lin$^{1,2,5,*}$,
  Tianyi Bai$^{3,*}$,
  Jiajun Zhang$^{4}$,
  Xuanhao Chang$^{2}$,
  Sheng Lu$^{2}$,
  Fangming Gu$^{2}$,
  Zengjie Hu$^{1,5}$,
  Wentao Zhang$^{1,\dagger}$\;\\
  $^{1}$PKU\;
  $^{2}$JLU\;
  $^{3}$HKUST\;
  $^{4}$USTC\;
  $^{5}$Shanghai AI Lab\;\\
  $^{*}$Equal contribution.\;
  $^{\dagger}$Corresponding author.
}

\date{}

\begin{document}
\maketitle
%\begingroup
%\renewcommand\thefootnote{}

%\addtocounter{footnote}{-1}
%\endgroup

\begin{abstract}
    Facial Expression Recognition (FER) is a fine-grained visual understanding task where reliable predictions require reasoning over localized and meaningful facial cues. Recent vision--language models (VLMs) enable natural language explanations for FER, but their reasoning is often ungrounded, producing fluent yet unverifiable rationales that are weakly tied to visual evidence and prone to hallucination, leading to poor robustness across different datasets. We propose TAG (Thinking with Action Unit Grounding), a vision--language framework that explicitly constrains multimodal reasoning to be supported by facial Action Units (AUs). TAG requires intermediate reasoning steps to be grounded in AU-related facial regions, yielding predictions accompanied by verifiable visual evidence. The model is trained via supervised fine-tuning on AU-grounded reasoning traces followed by reinforcement learning with an AU-aware reward that aligns predicted regions with external AU detectors. Evaluated on RAF-DB, FERPlus, and AffectNet, TAG consistently outperforms strong open-source and closed-source VLM baselines while simultaneously improving visual faithfulness. Ablation and preference studies further show that AU-grounded rewards stabilize reasoning and mitigate hallucination, demonstrating the importance of structured grounded intermediate representations for trustworthy multimodal reasoning in FER. The code will be available at \url{https://github.com/would1920/FER_TAG}.
\end{abstract}

\section{Introduction}
Facial Expression Recognition (FER) is a long-standing problem in computer vision and affective computing, with applications ranging from human--computer interaction to mental health analysis\cite{Ullah2024survey,sajjad2023survey, ghazouani2023survey, adyapady2023survey, martinez2016survey}. Despite significant progress enabled by deep neural networks\cite{He2015resnet,howard2017mobilenet,liu2021swintransformer, dosovitskiy2020visiontransformer}, modern FER systems largely operate as black boxes: they output an expression label without providing reliable evidence for why a particular prediction is made\cite{zhang2022eac,xue2022apvit, wang2020scn}. In high-stakes or real-world scenarios, such opacity limits trust and hinders deployment, motivating increasing interest in interpretable and explainable FER models.

Recent advances in vision\mbox{–}language models (VLMs)\cite{radford2021clip,liu2023llava,Qwen-VL,chen2024internvl, bai2025hallucination} offer a promising direction toward interpretable FER by enabling models to generate natural language explanations alongside predictions\cite{lan2025expllm,li2024cliper,zhang2025unifer}. Compared to traditional classifiers, VLM based approaches appear to reason explicitly about facial cues, suggesting a path toward human-understandable decision processes. However, this promise comes with a critical limitation. In practice, the explanations produced by general-purpose VLMs are often ungrounded: they are fluent and plausible, yet weakly tied to the actual visual evidence in the face image. Such models tend to behave more like storytellers than clinicians---producing convincing narratives without verifiable support from localized facial observations.

This lack of visual grounding leads to several fundamental problems. First, ungrounded explanations cannot be externally verified, undermining their usefulness for interpretability and trust. Second, the absence of explicit visual anchors encourages hallucination and shortcut reasoning, where models rely on dataset biases or high level semantic patterns rather than genuine facial evidence. Finally, these issues are exacerbated when models are evaluated across different datasets, where shifts in data distribution expose the brittleness of explanations that are not grounded in stable, physiologically meaningful cues. Together, these limitations reveal a key gap in current FER research: how to enable multimodal reasoning that is not only expressive, but also faithful to visual evidence.

We argue that FER provides a uniquely suitable testbed for addressing this problem. Human interpretation of facial expressions is inherently grounded in facial Action Units (AUs), as formalized in the Facial Action Coding System (FACS)\cite{ekman1978facs}. AUs correspond to localized and physiologically meaningful muscle activations, offering a structured intermediate representation between raw pixels and semantic emotion labels\cite{yang2019aufer, lien1998aufer}. Unlike free form textual rationales, AU related evidence can be explicitly localized on the face and independently verified using external detectors. This makes AU grounding a natural mechanism for constraining multimodal reasoning in FER.

Based on this insight, we propose TAG (Thinking with Action Unit Grounding), a vision\mbox{-}language framework that explicitly enforces grounded reasoning for FER. Instead of allowing free-form explanations, TAG requires the model to anchor its intermediate reasoning steps to AU-related facial regions, producing predictions accompanied by verifiable visual evidence. Notably, TAG utilizes pseudo-labels from SOTA AU detectors, which are themselves pretrained on expert-annotated data, to provide scalable and physiologically-aligned supervision without requiring additional manual labeling. TAG is trained in two stages: supervised fine-tuning on AU-grounded reasoning traces that teach the model how to reason with physiologically meaningful cues, followed by reinforcement learning with an AU-aware reward that aligns predicted regions with external AU detectors. This design ensures that improvements in recognition accuracy are achieved through faithful visual reasoning rather than ungrounded shortcuts.

We evaluate TAG on three widely used FER benchmarks, RAF-DB, FERPlus, and AffectNet. Extensive experiments show that TAG not only improves recognition accuracy compared to strong open\mbox{-}source and closed\mbox{-}source VLM baselines, but also substantially enhances visual faithfulness and robustness. Through ablation studies, training dynamics analysis, and human and LLM-based evaluations, we further demonstrate that unconstrained reinforcement learning can degrade visual grounding, while AU-aware rewards stabilize and strengthen physiologically aligned reasoning.

In summary, our main contributions are as follows:
\begin{itemize}
    \item We identify \emph{ungrounded multimodal reasoning} as a key limitation of existing vision--language approaches to facial expression recognition, where fluent explanations are weakly tied to visual evidence, leading to hallucination and poor robustness.
    \item We propose \textbf{TAG} (\textbf{T}hinking with \textbf{A}ction Unit \textbf{G}rounding), a physiologically grounded vision--language framework that explicitly constrains reasoning to be supported by facial Action Units via structured supervision and AU-aware reinforcement learning.
    \item We construct \textbf{TAG-310k}, a large-scale dataset of AU-grounded reasoning traces built on standard FER benchmarks, enabling supervised and reinforcement learning of faithful multimodal reasoning.
    \item We conduct extensive experiments on three FER benchmarks, including extensive ablation studies, which systematically validate the effectiveness of AU grounding for improving both recognition accuracy and visual faithfulness.
\end{itemize}

\begin{figure*}[t]
    \centering
    \includegraphics[width=\textwidth]{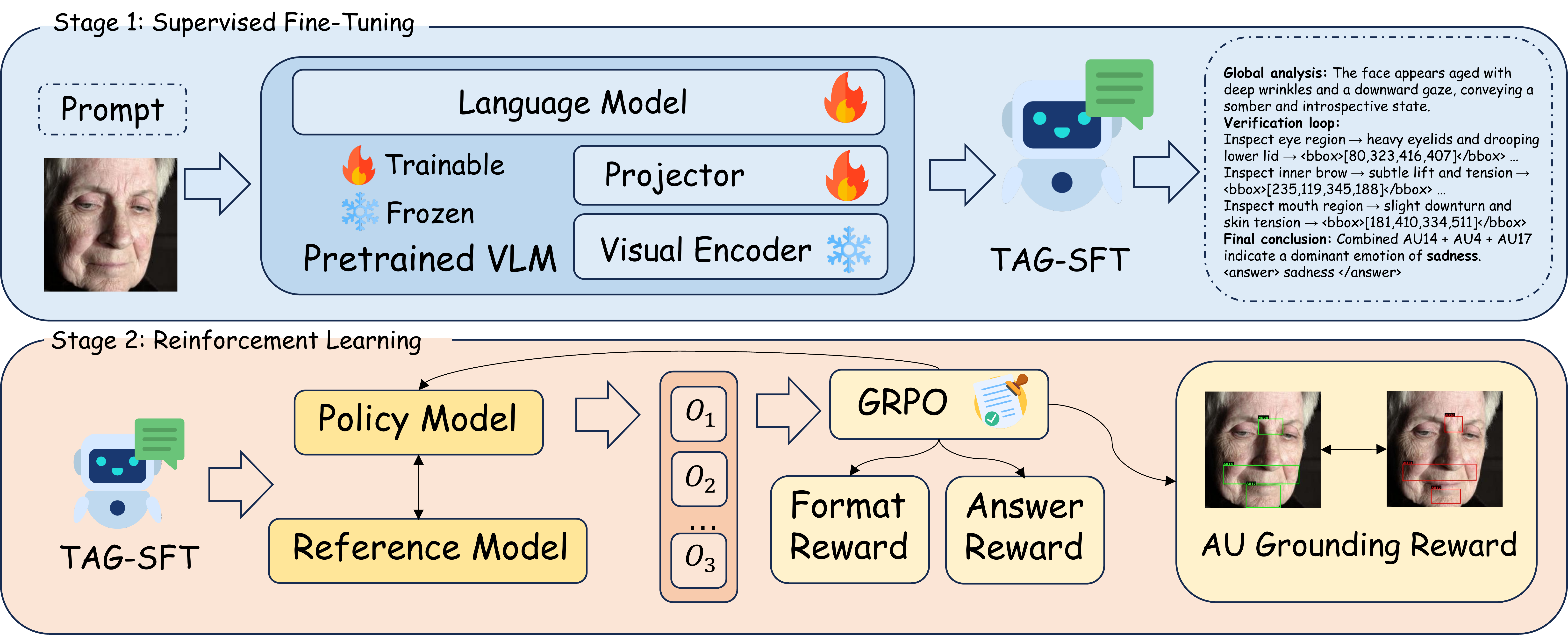}
    \caption{Two-stage training framework of TAG. TAG is first trained by supervised fine-tuning with AU-grounded reasoning traces, and then refined by GRPO using format, answer, and AU grounding rewards.}
    \label{fig:tag-qualitative}
\end{figure*}

\section{Related Work}
\label{sec:related}
\paragraph{Facial Expression Recognition.} Deep learning has significantly advanced Facial Expression Recognition (FER), driven by the growth of in-the-wild datasets~\cite{li2017rafdb,BarsoumICMI2016ferplus,mollahosseini2017affectnet} and robust visual backbones. While the advent of CLIP facilitated vision-language alignment like CLIPER and Exp-CLIP~\cite{zhao2025expclip}, both traditional classifiers and contrastive methods remain "black boxes" lacking interpretability. Recently, Multimodal Large Language Models (MLLMs) have shifted the paradigm toward explainable FER. Notably, methods like ExpLLM\cite{lan2025expllm} and UniFER\cite{zhang2025unifer} have begun constructing Action Unit datasets to empower models with reasoning capabilities, moving beyond simple classification.

\paragraph{Multimodal Large Language Models. } The integration of Large Language Models (LLMs) with visual encoders has catalyzed the development of Multimodal Large Language Models (MLLMs). Representative works such as LLaVA, Qwen-VL, and Intern-VL leverage massive image-text pairs and instruction tuning to endow LLMs with robust visual perception capabilities. These models excel at open-ended visual reasoning and instruction following, enabling them to handle diverse vision-language tasks. Most recently, the emergence of models like GPT-o3~\cite{openai2025gpto3}, DeepEyes~\cite{zheng2025deepeyes} and GRIT~\cite{fan2025grit} have introduced "thinking with images" as a new paradigm for multimodal reasoning. 

\section{Method: TAG}
\label{sec:method}

TAG (Thinking with Action Unit Grounding) is a vision-language framework that explicitly constrains the VLM's reasoning process to be supported by physiologically meaningful AUs. Instead of predicting a label from a global image feature, TAG must (i) reason explicitly, (ii) ground intermediate steps to AU-related regions, and (iii) output a final FER label together with verifiable visual evidence.

\subsection{Overview and Objective}
\label{sec:overview}

\paragraph{Architecture.}
TAG follows a standard VLM design: a visual encoder $f_{\mathrm{v}}$ maps an input face image $x$ to visual tokens $v = f_{\mathrm{v}}(x)$, and a language model $f_{\mathrm{llm}}$ autoregressively generates tokens $w = (w_1, \dots, w_T)$ conditioned on $v$. The model directly learns to emit up to $M$ bounding boxes $\mathcal{B} = \{b_i\}_{i=1}^{M}$, each as normalized coordinates $(x_1^i, y_1^i, x_2^i, y_2^i)$, together with the final FER label $y \in \mathcal{Y}$.

\paragraph{TAG Objective.}
At a high level, TAG learns a mapping from images to AU-grounded explanations, $f_{\theta}: x \mapsto (\hat{y}, \hat{\mathcal{B}}, \hat{\tau})$, where $\hat{y}$ is the predicted expression, $\hat{\mathcal{B}}$ are supporting boxes, and $\hat{\tau}$ is the reasoning trace. The goal is that $\hat{y}$ matches the FER label and that $\hat{\mathcal{B}}$/$\hat{\tau}$ stay consistent with externally detected AUs. The two-stage training (SFT + RL) is arranged so that AU grounding and reasoning reinforce each other rather than allowing ungrounded shortcuts.

\subsection{AU-Grounded Thinking via Supervised Fine-Tuning}
\label{sec:sft}

\paragraph{Structured AU-Grounded Thinking.}
We cold-start TAG with supervised fine-tuning (SFT) on demonstrations written in a fixed natural-language format: a short \texttt{<think>} block that narrates global-to-local reasoning, contains up to three \texttt{<bbox>}[x\textsubscript{1}, y\textsubscript{1}, x\textsubscript{2}, y\textsubscript{2}]\texttt{</bbox>} references to AU-related regions, and ends with an \texttt{<answer>} label. Each example includes the image, detected AU cues, and this reasoning trace. Unlike cropping-based methods, TAG keeps a single global visual encoding $v$ for the whole face image and does not modify the attention mechanism or perform any region-level visual operations.
The <bbox> tokens are simply predicted by the decoder as part of the text sequence, and spatial grounding is learned through supervision rather than enforced by architectural design.

\paragraph{Supervised Objective.}
SFT maximizes the likelihood of the structured target sequence:

\begin{equation}
\label{eq:sft-loss}
\mathcal{L}_{\mathrm{SFT}} = - \sum_{t} \log p_{\theta}(w_t \mid w_{<t}, f_{\mathrm{v}}(x)),
\end{equation}

where $w_t$ spans the reasoning text, box coordinates, and the final label. This teaches TAG to produce AU-grounded traces whose intermediate evidence can later be checked by independent AU detectors and prepares the model for AU-aware rewards in RL.

\subsection{RL with AU-Grounded Reward}
\label{sec:rl}

\paragraph{GRPO Optimization.}
Although SFT produces reasonable AU-grounded behavior, the model still inherits hallucination from generic VLM pretraining. We therefore apply GRPO~\cite{shao2024grpo} on top of the SFT checkpoint. Following the standard GRPO formulation, we optimize the policy using:

\begin{equation}
    \label{eq:grpo}
    \mathcal{L}_{\mathrm{GRPO}}(\theta)
    = \mathbb{E}_{O_i \in \mathcal{G}}
    \Bigg[
    \frac{1}{G}
    \sum_{i=1}^G
    \min \Big(
     \rho_i A_i, \\
    \mathrm{clip}\!\left(
    \rho_i, 1 - \epsilon, 1 + \epsilon
    \right) A_i
    \Big)
    \Bigg]
    - \beta \,
    D_{\mathrm{KL}}
    \!\left(
    \pi_\theta \,\|\, \pi_{\theta_{\mathrm{old}}}
    \right).
    \end{equation}
    
where $A_i = \frac{R_i - \mathrm{mean}(\{R_j\})}{\mathrm{std}(\{R_j\})}$ is the normalized advantage, $\rho_i = \frac{\pi_\theta(O_i)}{\pi_{\theta_{\mathrm{old}}}(O_i)}$ is the importance sampling ratio, $\mathcal{G}$ is a group of $G$ trajectories, $\epsilon$ is the clipping parameter, and $\beta$ is the KL penalty coefficient. Here, $\pi_{\theta_{\mathrm{old}}}$ is the reference policy.

\begin{figure*}[t]
    \centering
    \includegraphics[width=\textwidth]{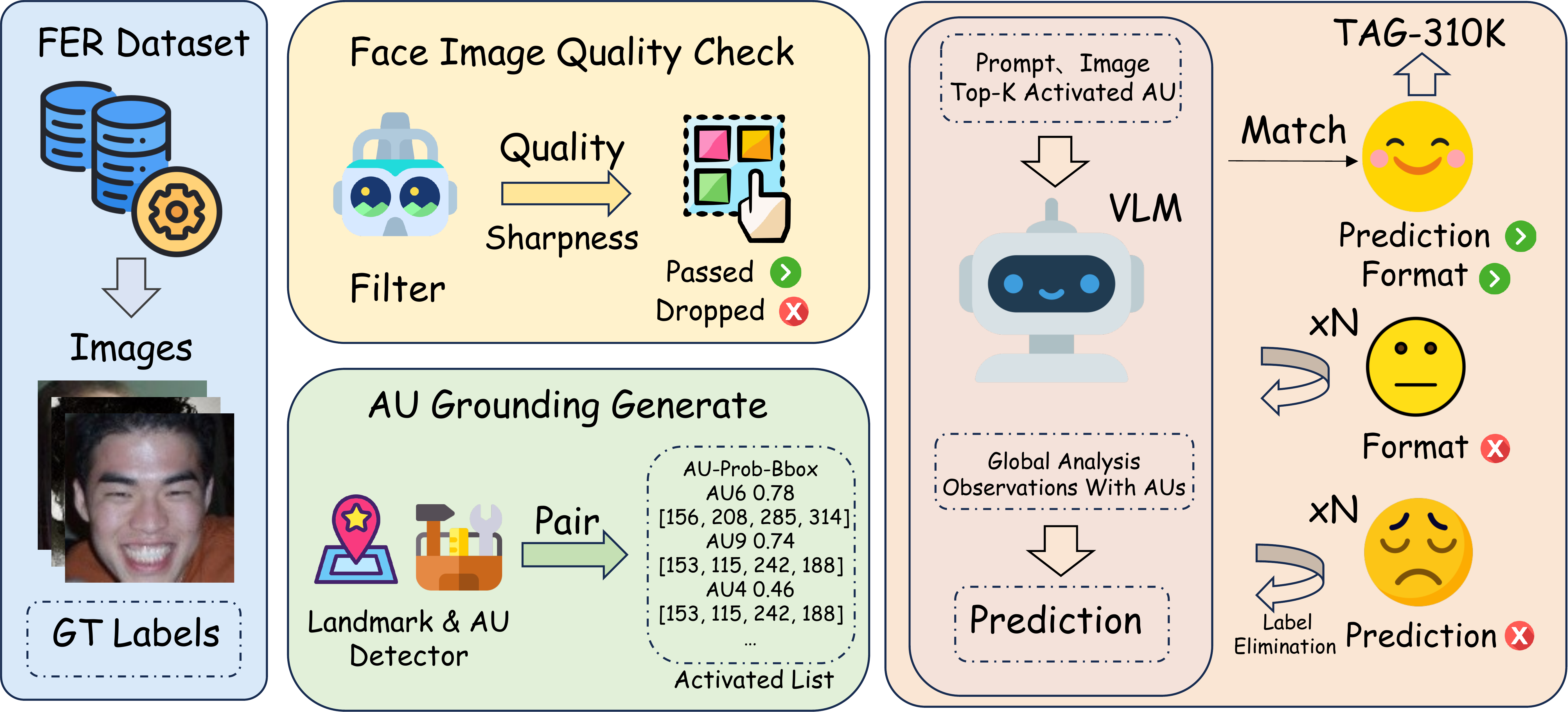}
    \caption{Construction pipeline of TAG-310k. Face images are filtered, annotated with activated AUs, and used to generate and select high-quality AU-grounded reasoning trajectories.}
    \label{fig:tag-data}
\end{figure*}

\paragraph{AU-IoU Reward.}
The core of TAG is an AU-grounded reward that measures how well predicted boxes align with external AU detectors. Let $\mathcal{B}_{\mathrm{pred}} = \{b_i\}_{i=1}^{M}$ be the boxes decoded in a trajectory and $\mathcal{B}_{\mathrm{AU}}^{\mathrm{act}} = \{\hat{b}_j\}_{j=1}^{L}$ the boxes of activated AUs produced by an AU detector. For each $b_i$, we compute its IoU with the closest AU box:

\begin{equation}
\label{eq:au-iou-single}
\mathrm{IoU}(b_i) = \max_{j} \frac{\mathrm{area}(b_i \cap \hat{b}_j)}{\mathrm{area}(b_i \cup \hat{b}_j)}.
\end{equation}

Rather than requiring strict one-to-one matching between predicted boxes and specific AU types, we use the maximum IoU across all activated AUs. This design mitigates noise from external AU detectors and guides the model to attend to relevant facial regions, improving robustness. For AU reward, we consider only samples with activated AU ground-truth boxes; others skip AU-gradient. Suppose $k=|\mathcal{B}_{\mathrm{pred}}|$ predicted boxes and $n=|\mathcal{B}_{\mathrm{AU}}^{\mathrm{act}}|$ activated AU boxes. For each predicted $b_i$, we first find the AU box $\hat{b}_j$ with maximum IoU (i.e., $\arg\max_j \mathrm{IoU}(b_i, \hat{b}_j)$), then compute $\mathrm{IoU}(b_i)$ as defined in Equation~\eqref{eq:au-iou-single}. To prevent the model from hacking the reward by predicting excessive boxes, we average only the top-$\min(n,k)$ IoU values:

\begin{equation}
\label{eq:au-iou-avg}
R_{\mathrm{AU}} = \dfrac{1}{\min(n,k)} \sum_{i=1}^{\min(n,k)} \mathrm{IoU}_{(i)},
\end{equation}

where $\mathrm{IoU}_{(i)}$ is the $i$-th largest IoU among the $k$ predicted boxes. This ensures that when $k > n$ (e.g., $k=3$ predicted boxes but only $n=2$ activated AUs), only the best $\min(n,k)$ matches contribute to the reward, preventing reward hacking through over-prediction.

\paragraph{Answer and Format Rewards.}
We complement $R_{\mathrm{AU}}$ with two lightweight signals. The answer reward $R_{\mathrm{ans}}$ is $1$ if the final label in \texttt{<answer>} matches the ground-truth FER label, and $0$ otherwise. The format reward $R_{\mathrm{fmt}}$ is defined as: a single \texttt{<think>}--\texttt{</think>} block, at most three \texttt{<bbox>} entries inside it, and exactly one \texttt{<answer>} span with a valid label; well-formed outputs receive a small positive bonus, malformed ones receive no bonus.

\paragraph{Overall Reward.}
For each sampled trajectory we sum the three components into a single scalar:

\begin{equation}
\label{eq:total-reward}
R = R_{\mathrm{AU}} + R_{\mathrm{ans}} + R_{\mathrm{fmt}}.
\end{equation}

GRPO then updates the policy toward trajectories with larger $R$, reducing reward sparsity compared to pure answer-only RL and explicitly penalizing hallucinated or ungrounded visual evidence.

\section{AU-Grounded Reasoning Data}
\label{sec:data}

\paragraph{Data Sources.}
We construct TAG-310K by aggregating samples from three widely used facial expression recognition benchmarks:
AffectNet~\cite{mollahosseini2017affectnet}, FERPlus~\cite{BarsoumICMI2016ferplus}, and RAF-DB~\cite{li2017rafdb}.
To avoid any potential data leakage, we strictly use only the official training splits of all datasets.
Each face image is processed by an AU detector (GraphAU~\cite{luo2022graphau}) and a landmark detector (MediaPipe~\cite{lugaresi2019mediapipe}).
Detected AUs are mapped to their corresponding landmark regions, yielding a set of AU bounding boxes
$\mathcal{B}_{\mathrm{det}}$, which are later used for AU-grounded reasoning and quality filtering.

\paragraph{Filtering and AU Grounding Preparation.}
We adopt a multi-stage filtering strategy to ensure visual quality and reliable AU grounding.
First, a vision--language model (VLM) is used to perform automatic image quality assessment;
samples with severe blur, occlusion, extreme pose, or missing facial regions are discarded.
For the remaining images, we extract the top-$K$ activated AUs and their corresponding bounding boxes
from $\mathcal{B}_{\mathrm{det}}$, which serve as structured cues for subsequent reasoning trace generation.
This stage ensures that all retained samples are associated with explicit and localized facial action evidence.

\paragraph{Reasoning Trace Generation.}
For each qualified sample, we prompt an open-source VLM with the image and the detected AUs to generate
a structured reasoning trace in a fixed
\texttt{<think>}/\texttt{<bbox>}/\texttt{<answer>} format.
The reasoning trace is required to consist of two parts:
(i) a \emph{global observation} describing holistic facial appearance and affective cues, and
(ii) \emph{local AU grounding} that explicitly links specific AUs to localized facial regions via bounding boxes.

During generation, we consider three possible cases.
\textbf{(1) Correct format and correct prediction:}
if both the output format and the predicted expression label match the ground-truth annotation,
the sample is directly accepted into the dataset.
\textbf{(2) Format error:}
if the model violates the required output format, we retry generation with the same inputs until a valid format is obtained.
\textbf{(3) Prediction mismatch:}
if the format is correct but the predicted label does not match the ground truth,
we do not directly provide the correct label to the model.
Instead, we adopt an \emph{iterative label elimination} strategy:
the incorrect predicted label is removed from the candidate label set,
and the model is prompted again to select and justify an expression from the remaining labels.
This process is repeated until the correct label is obtained or the candidate set is exhausted.
In this way, we encourage the model to refine its reasoning by exclusion rather than relying on explicit label supervision,
leading to more faithful AU-grounded explanations.

Finally, we perform random human inspection on a subset of generated samples to verify
the correctness of both expression labels and AU-based reasoning consistency.

Detailed prompt templates and pipeline implementations are provided in
Appendix~\ref{sec:appendix-a} and Appendix~\ref{sec:appendix-b}.

\section{Experiments}
\label{sec:exp}
\subsection{Implementation Details}
We train TAG using a two-stage approach: supervised fine-tuning on TAG-310k followed by per-dataset reinforcement learning. For TAG-310k construction, we use Qwen2.5-VL-32B for filtering and reasoning trace generation. As a base model for TAG training, we use Qwen2.5-VL-7B and freeze the visual encoder while unfreezing projector and decoder so that both visual perception and language reasoning can adapt to AU-grounded supervision. All face images are resized to $224\times224$ (with aspect ratio preserved) as input to the vision encoder.
Bounding box coordinates in autoregressive grounding prediction are instead normalized with respect to a virtual $512\times512$ canvas, which defines the coordinate system for spatial supervision.

\begin{table*}[t]
    \centering
    \small
    % 缩小列间距，以便容纳完整的长模型名称
    \setlength{\tabcolsep}{10pt} 
    
    \caption{Main results (accuracy \%) on RAF-DB, FERPlus, and AffectNet. \textbf{Bold} indicates the best performance in each column, \underline{underlined} indicates the second-best. $^{\dagger}$Methods with this marker use per-dataset tuning (one model per dataset). TAG-7B (SFT only) uses a single unified model for all datasets.}
    \label{tab:main}
    
    \begin{tabular}{lcccc}
        \toprule
        Model & RAF-DB & FERPlus & AffectNet & Average \\
        \midrule
        \multicolumn{5}{l}{\textit{Open-source VLMs, Zero Shot}} \\
        LLaVA-Next-Llama3-8B~\cite{zhang2024llavanext} & 59.09 & 73.36 & 37.17 & 56.54 \\
        Qwen2.5-VL-7B~\cite{bai2025qwen2.5vl} & 56.75 & 66.36 & 33.89 & 52.33 \\
        InternVL3.5-8B~\cite{zhu2025internvl3} & 73.47 & 69.41 & 38.57 & 60.48 \\
        LLaVA-Next-34B~\cite{zhang2024llavanext} & 78.36 & 69.38 & 42.83 & 63.52 \\
        InternVL3-38B~\cite{zhu2025internvl3} & 79.95 & 68.75 & 46.63 & 65.11 \\
        \midrule
        \multicolumn{5}{l}{\textit{Closed-source VLMs, Zero Shot}} \\
        Gemini-2.5-Pro~\cite{google2025gemini25} & 65.51 & 56.33 & 48.49 & 56.78 \\
        Gemini-2.5-Flash~\cite{google2025gemini25} & 68.84 & 70.06 & 50.14 & 63.01 \\
        GPT-5~\cite{openai2025gpt5} & 74.05 & 67.84 & 46.91 & 62.93 \\
        \midrule
        \multicolumn{5}{l}{\textit{FER-specific Methods, Fine-Tuned}} \\
        SCN$^{\dagger}$~\cite{wang2020scn} & 87.03 & 88.01 & -- & -- \\
        EAC$^{\dagger}$~\cite{zhang2022eac} & 89.99 & 89.64 & 65.32 & 81.65 \\
        APViT$^{\dagger}$~\cite{xue2022apvit} & 91.98 & 90.86 & 66.91 & 83.25 \\
        POSTER$^{\dagger}$~\cite{zheng2023poster} & \underline{92.05} & \textbf{91.62} & \textbf{67.31} & 83.66 \\
        ExpLLM$^{\dagger}$~\cite{lan2025expllm} & 91.03 & 90.50 & 65.93 & 82.49 \\
        UniFER~\cite{zhang2025unifer} & 88.72 & 76.49 & 48.50 & 71.24 \\
        \midrule
        \multicolumn{5}{l}{\textit{Our Method}} \\
        \rowcolor{gray!10}
        TAG-7B (SFT only) & 89.02 & 81.01 & 50.00 & 74.34 \\
        \rowcolor{gray!25}
        \textbf{TAG-7B (ours, RL)$^{\dagger}$} & \textbf{92.80} & \underline{91.50} & \underline{67.03} & \textbf{83.78} \\
        \bottomrule
    \end{tabular}
\end{table*}

\paragraph{Supervised Fine-Tuning.}
For the SFT stage, we train on TAG-310k (310k AU-grounded samples) using AdamW~\cite{loshchilov2017adamw} with learning rate $2\!\times\!10^{-5}$, cosine decay, and standard weight decay. We use a global batch size chosen to fit 4$\times$H200 GPUs and stop when validation performance saturates. This unified SFT stage enables the model to learn AU-grounded reasoning across all datasets simultaneously.

\paragraph{Reinforcement Learning.}
For the RL stage, we apply GRPO on top of the SFT checkpoint. Unlike SFT which uses the full TAG-310k, RL is performed separately for each dataset using the corresponding subset of TAG-310k (e.g., RAF-DB subset for RAF-DB evaluation). For each dataset-specific RL, we use rollout number $K\!=\!8$, KL penalty coefficient $0.04$, and learning rate $1\!\times\!10^{-6}$ for 2 epochs. At inference time we set the decoding temperature to 0 and restrict the model to at most three \texttt{<bbox>} tokens inside \texttt{<think>}, based on the analysis of AU selection patterns in TAG-310k (see Appendix~\ref{sec:appendix-a12}).

\subsection{Main Results}
\label{sec:main-results}
Table~\ref{tab:main} presents comprehensive results comparing TAG against open-source and closed-source VLMs, as well as FER-specific methods. We evaluate TAG under two settings: (1) SFT-only model trained on TAG-310k and evaluated uniformly across all datasets, and (2) RL model with per-dataset tuning, where RL is performed separately on each dataset-specific subset of TAG-310k and evaluated on the corresponding dataset.

\begin{figure*}[t]
    \centering
    \includegraphics[width=\textwidth]{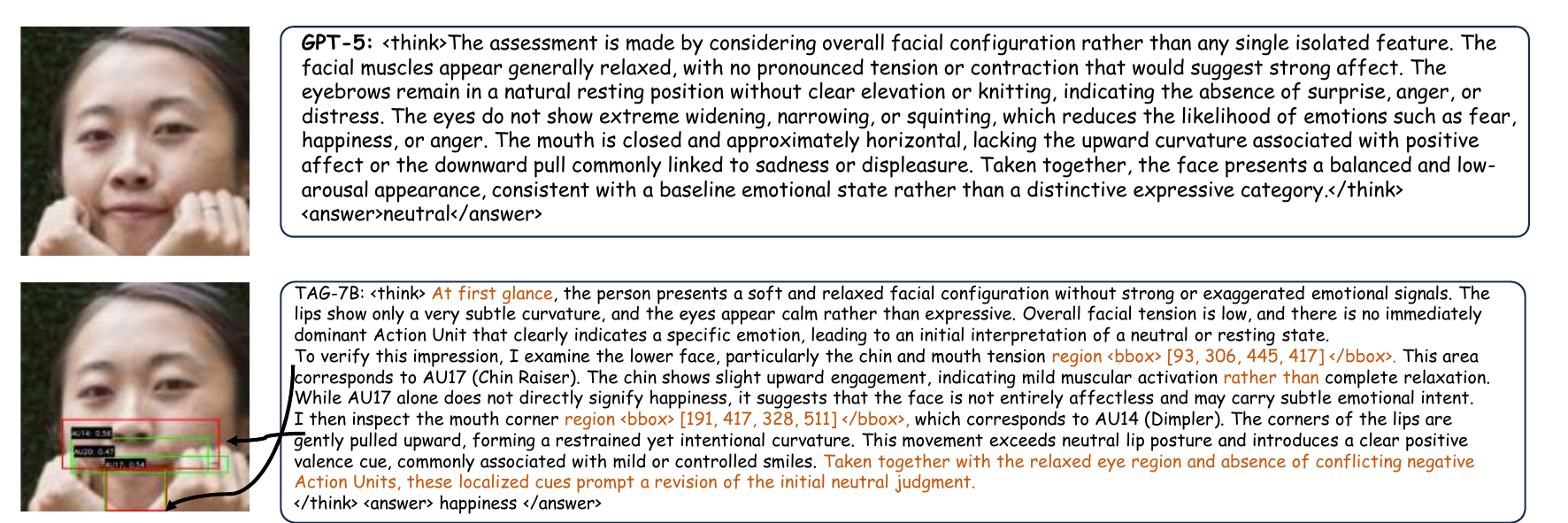}
    \caption{Qualitative case studies of TAG's AU-grounded reasoning. Green boxes indicate external AU detector outputs with confidence scores, while red boxes represent TAG's predicted regions.}
    \label{fig:case-study}
\end{figure*}

\paragraph{TAG Achieves SOTA in Uniform Setting.}
TAG achieves 74.34\% average accuracy across RAF-DB, FERPlus, and AffectNet using a single model trained with supervised fine-tuning, significantly outperforming open-source VLMs by +9.23 points over InternVL3-38B and closed-source models by +11.41 points over GPT-5, despite using a much smaller 7B backbone. This demonstrates that AU-grounded supervision enables strong generalization.

\paragraph{TAG Achieves SOTA in Per-Dataset Setting.}
With per-dataset reinforcement learning tuning, TAG achieves state-of-the-art performance on individual benchmarks: 92.80\% on RAF-DB, 91.50\% on FERPlus, and 67.03\% on AffectNet, with an average of 83.78\%. This surpasses all FER-specific methods including SCN, EAC, APViT, POSTER and ExpLLM, demonstrating that AU-grounded reinforcement learning further enhances performance when dataset-specific tuning is applied.

\subsection{Ablation Study}
\label{sec:ablation}
We ablate the two-stage training scheme and AU-grounded reward on RAF-DB. Specifically, we compare: (1) the base VLM Qwen2.5-VL-7B without AU grounding, (2) adding AU-grounded SFT only, (3) further adding RL with $R_{\mathrm{fmt}}$ and $R_{\mathrm{ans}}$ rewards only (denoted as RLVR), and (4) our full TAG with AU-grounded reward ($R_{\mathrm{fmt}}$ + $R_{\mathrm{ans}}$ + $R_{\mathrm{AU}}$). We track FER accuracy and AU IoU between predicted boxes and external AU detectors. Additionally, Appendix~\ref{appendix:dynamics} visualizes the training dynamics during both SFT and RL stages, providing insights into how TAG learns AU-grounded reasoning.

\paragraph{SFT Provides the Foundation for Performance.}
\begin{table}[t]
\centering
\small
\begin{tabular}{cccccc}
\toprule
SFT & $R_{\mathrm{ans}}, R_{\mathrm{fmt}}$ & $R_{\mathrm{AU}}$ & Acc & IoU \\
\midrule
 &  &  & 56.75 & -- \\
$\checkmark$ &  &  & 89.02 & 46.73 \\
$\checkmark$ & $\checkmark$ &  & 90.22 & 43.46 \\
$\checkmark$ & $\checkmark$ & $\checkmark$ & \textbf{92.80} & \textbf{60.24} \\
\bottomrule
\end{tabular}
\caption{Ablation on TAG components with cumulative checkmarked modules and RAF-DB accuracy / AU IoU. The final row shows per-dataset tuning results.}
\label{tab:ablation}
\end{table}

Table~\ref{tab:ablation} shows that AU-grounded SFT establishes the foundation for TAG's performance. Teaching the model to follow structured AU-grounded reasoning traces lifts accuracy from $56.75$ to $89.02$ and yields a strong IoU of $46.73$. This improvement indicates that learning to reason with AU evidence through supervised demonstrations enables the model to attend to physiologically meaningful facial regions and generate structured reasoning traces.

\paragraph{RLVR May Degrade Visual Grounding.}
Adding RL with only $R_{\mathrm{fmt}}$ and $R_{\mathrm{ans}}$ rewards brings an accuracy gain but degrades AU IoU from $46.73$ to $43.46$. This reveals an important trade-off: while unconstrained RL can improve task-specific accuracy by optimizing directly for correct labels, it may inadvertently hurt the model's visual grounding capability. Without explicit AU supervision in the reward function, the RL optimization tends to exploit shortcut patterns and drift toward less faithful visual alignments.

\paragraph{TAG with AU-Grounded Reward Improves Both Accuracy and Visual Grounding.}
When we introduce the AU-IoU reward in our full TAG framework, the model simultaneously improves both accuracy and visual grounding. With AU-grounded RL, TAG achieves $92.80$ Acc and $60.24$ IoU on RAF-DB, demonstrating that the proposed AU-grounded reward not only prevents the visual degradation observed with generic RL but actually enhances visual grounding substantially beyond the SFT baseline. Overall, the ablations support our claim that AU-grounded SFT provides a strong initialization, and AU-aware RL is necessary to further improve performance while maintaining and even enhancing visual grounding capabilities.

\subsection{Mechanism Analysis}
\label{sec:mechanism}
We conduct further analysis to understand the mechanisms underlying TAG's performance. Specifically, we investigate (1) whether grounding itself brings benefits beyond structured reasoning, and (2) how different reward designs affect model behavior.

\paragraph{Grounding Benefits in SFT.}
To investigate whether grounding itself brings benefits, we conduct an ablation study on the SFT stage. Specifically, we compare three settings: (1) baseline without AU grounding (Qwen2.5-VL-7B zero-shot), (2) SFT with all bounding-box spans removed (i.e., deleting the entire content enclosed by \texttt{<bbox>} and \texttt{</bbox>} while keeping all other tokens and structure identical), and (3) full SFT with complete AU-grounded reasoning traces. This design allows us to isolate the contribution of visual grounding from the structured reasoning format.

\begin{table}[t]
\centering
\small
\setlength{\tabcolsep}{3.5pt}
\begin{tabular}{lccc}
\toprule
Method & RAF-DB & FERPlus & AffectNet \\
\midrule
Baseline (w/o grounding) & 56.75 & 66.36 & 33.89 \\
SFT w/o \texttt{<bbox>} & 86.99 & 78.99 & 48.09 \\
SFT (full) & 89.02 & 81.01 & 50.00 \\
\midrule
$\Delta$ (w/o \texttt{<bbox>} vs full) & -2.03 & -2.02 & -1.91 \\
\bottomrule
\end{tabular}
\caption{Ablation on grounding capability in SFT stage. We remove all \texttt{<bbox>} content while keeping other structure identical to evaluate grounding benefits.}
\label{tab:grounding}
\end{table}

Table~\ref{tab:grounding} shows that grounding provides substantial benefits. Removing bounding-box spans from the SFT training leads to a performance drop of $1.98$ points on average ($-2.03$ on RAF-DB, $-2.02$ on FERPlus, $-1.91$ on AffectNet) compared to the full SFT model. This demonstrates that the explicit visual grounding through bounding box annotations is crucial for the model's performance, beyond the benefits of structured reasoning format alone. The grounding mechanism enables the model to attend to physiologically meaningful facial regions, which in turn improves expression recognition accuracy.

\paragraph{Reward Design and Cross-Detector Evaluation.}

As shown in Section~\ref{sec:ablation}, incorporating the AU-grounded reward $R_{\mathrm{AU}}$ improves both accuracy and grounding quality. 
However, since AU supervision relies on external detectors, a potential concern is whether grounding improvements are detector-specific. 
To examine whether AU-grounded learning generalizes beyond the training detector, we evaluate grounding using AU regions derived from an independent detector OpenFace\cite{Hu2025openface}, which is not used during data construction or reward computation.

\begin{table}[t]
\centering
\small
\setlength{\tabcolsep}{7pt}
\begin{tabular}{lcccc}
\toprule
Reward Design & \multicolumn{2}{c}{RAF-DB} & \multicolumn{2}{c}{FERPlus} \\
 & Acc & IoU & Acc & IoU \\
\midrule
IoU-based & 92.80 & 61.78 & 91.50 & 53.27 \\
F1-based & 92.31 & 56.46 & 91.21 & 50.84 \\
\bottomrule
\end{tabular}
\caption{Comparison of AU reward designs: IoU-based (validating bounding box coordinates against AU regions derived from an independent detector, OpenFace) vs F1-based (validating extracted AU labels).}
\label{tab:au-reward}
\end{table}

We compare two AU reward formulations: an IoU-based reward that validates spatial alignment with AU regions, and an F1-based reward that validates only AU label consistency. 
AU labels are extracted from model responses using regular expressions, and F1-score against TAG-310k annotations is used as $R_{\mathrm{AU}}$, while keeping $R_{\mathrm{fmt}}$ and $R_{\mathrm{ans}}$ unchanged. 
As shown in Table~\ref{tab:au-reward}, the IoU-based reward achieves higher accuracy and consistently better grounding under OpenFace-based evaluation 
(e.g., $61.78$ vs $56.46$ IoU on RAF-DB, and $53.27$ vs $50.84$ on FERPlus), indicating that spatial grounding provides stronger supervision than AU-label-only validation.

\subsection{Case Study: Qualitative Analysis}
\label{sec:case-study}
We qualitatively examine TAG's predictions by overlaying the generated \texttt{<bbox>} regions on input faces and visualizing the model's reasoning process. Figure~\ref{fig:case-study} shows representative examples where TAG demonstrates AU-grounded reasoning capabilities.

TAG learns to predict AU-related regions that closely align with external AU detection tools. By comparing TAG's predicted bounding boxes with the outputs from GraphAU, we observe high spatial agreement: TAG's boxes consistently center on meaningful facial regions that correspond to activated AUs. This demonstrates that TAG learns the physiological structure of facial action units through the AU-grounded training process, rather than memorizing facial regions. We also observe cases where TAG corrects initial misimpressions through AU-grounded refinement. The structured format encourages the model to move from global observations to fine-grained AU-based evidence, and the AU-IoU reward reinforces predictions that are supported by verifiable facial regions. More comprehensive analysis of grounding behavior, including consistency with external detectors and emergent capabilities beyond detector coverage, is provided in Appendix~\ref{sec:appendix-c}.

\subsection{Cross-Dataset Evaluation}
\label{sec:cross-dataset}
To evaluate the generalization capability of TAG, we conduct cross-dataset experiments where we start from the unified SFT model and then further train it with RL on one dataset while evaluating on others. Specifically, we perform RL training on either RAF-DB or FERPlus and then evaluate the resulting policy on all three datasets without any additional fine-tuning.

\begin{table}[t]
\centering
\small
\setlength{\tabcolsep}{3.5pt}
\begin{tabular}{lcccc}
\toprule
Training Dataset & RAF-DB & FERPlus & AffectNet & Avg. \\
\midrule
RL on RAF-DB & 92.80 & 82.30 & 51.50 & 75.53 \\
$\Delta$ (vs SFT, RAF-DB) & +3.78 & +1.29 & +1.50 & +1.19 \\
\midrule
RL on FERPlus & 89.50 & 91.50 & 50.51 & 77.17 \\
$\Delta$ (vs SFT, FERPlus) & +0.48 & +10.49 & +0.51 & +2.83 \\
\bottomrule
\end{tabular}
\caption{Cross-dataset evaluation: RL training on RAF-DB or FERPlus, evaluation on all three datasets.}
\label{tab:cross-dataset}
\end{table}

Table~\ref{tab:cross-dataset} shows that RL on RAF-DB improves accuracy over the unified SFT model on all three datasets, with the largest gain on RAF-DB itself (+3.78 points to 92.80\%) and smaller but consistent gains on FERPlus (+1.29) and AffectNet (+1.50). Training RL on FERPlus similarly yields a strong boost on FERPlus (+10.49 points to 91.50\%) and moderate improvements on RAF-DB (+0.48) and AffectNet (+0.51), indicating that policies learned on a single source dataset can still benefit other domains. Overall, RL on RAF-DB and FERPlus raises the average accuracy from 74.34\%/74.34\% to 75.53\%/77.17\%, enhancing both in-domain and cross-dataset performance, while Table~\ref{tab:main} indicates that fully closing the remaining domain gaps still benefits from per-dataset RL tailored to each target distribution.

\section{Conclusion}
In this work, we proposed TAG (Thinking with Action Unit Grounding), a vision–language framework that explicitly grounds facial expression reasoning in physiologically meaningful Action Units. By constraining intermediate reasoning steps to localized AU evidence and reinforcing this behavior through AU-aware rewards, TAG transforms FER from holistic appearance-based prediction into evidence-driven, verifiable reasoning. Built on the large-scale TAG-310K dataset of AU-grounded reasoning traces, our approach consistently outperforms substantially larger VLMs and strong FER-specific models, while delivering more faithful and interpretable predictions. Extensive ablations and preference studies further demonstrate that structured grounding is essential for stabilizing multimodal reasoning and mitigating hallucination. We believe TAG provides a principled paradigm for trustworthy fine-grained visual understanding, and opens new directions for grounding-based multimodal reasoning in affective computing and beyond.

\clearpage
\section*{Impact Statement}

This work uses publicly available facial expression datasets that are widely adopted in prior research.
We do not collect new personal data, perform identity recognition, or introduce annotations that identify individuals,
and all experiments follow the original dataset licenses and usage conditions.
To respect data ownership and privacy, we do not redistribute face images and instead release only derived annotations,
including structured reasoning traces, Action Unit (AU) detection results, and region-level grounding information.
Access to the original images should be obtained directly from the dataset providers.

We acknowledge that facial expression recognition may raise ethical concerns if applied without appropriate context,
such as in surveillance or sensitive behavioral monitoring scenarios.
Our work is intended for controlled research settings, with the goal of improving transparency and interpretability
by grounding model reasoning in physiologically meaningful and verifiable visual evidence.
We hope that promoting grounded and explainable decision processes can help mitigate risks associated with
ungrounded or misleading model predictions in affective computing applications.

\clearpage

\bibliographystyle{unsrt}
\bibliography{ref}

\clearpage
\appendix

\label{sec:appendix}
\section{Discussion}
\subsection{Contributions}
\paragraph{For Multimodal Large Language Models.}
Emotion understanding has long been a central yet challenging topic for multimodal large language models (MLLMs), especially for facial expression recognition (FER) where inter-class similarity is high and subtle muscular nuances often determine the correct label.
Our work investigates how to make MLLMs read facial expressions in a more faithful and reliable manner by grounding their reasoning process on explicit facial Action Units (AUs).
By explicitly tying each reasoning step to localized AU evidence, TAG transforms the FER task from ``guessing from holistic appearance'' into ``explaining from physiologically meaningful muscle activations.''
This AU-grounded formulation not only improves robustness on difficult cases, but also offers a generic recipe for steering MLLMs toward anatomically grounded, non-hallucinatory visual reasoning.

\paragraph{For Facial Expression Recognition.}
From the perspective of FER, our method substantially improves interpretability compared with conventional backbone+head architectures, which behave like black-box classifiers that output accurate predictions but provide little insight into why they are correct.
Recent MLLM-based FER systems such as ExpLLM and UniFER demonstrate that chain-of-thought (CoT) style reasoning can make FER more transparent, yet it remains unclear what constitutes a \emph{good} reasoning chain for FER and how the model should ``think'' about facial expressions.
Our TAG framework addresses this gap by proposing \emph{Thinking with Action Units Grounding}: the model is encouraged to explicitly select and reason over AU evidence, aligning its internal reasoning structure with the widely adopted Facial Action Coding System.
To the best of our knowledge, this is the first systematic application of AU-grounded, thinking-with-images style reasoning to FER, providing a new paradigm for interpretable affective computing.

\paragraph{Beyond FER: Future Directions.}
Our experiments show that grounding MLLM reasoning in AU evidence brings clear benefits on a fine-grained, high-similarity task such as FER.
We believe that this grounding-based perspective is not limited to facial expressions and can be extended to a broader class of fine-grained visual and affective tasks.
Future work could explore analogous grounding schemes for other structured physiological or semantic cues (e.g., body pose units, gaze patterns, or scene affordances), and investigate how to automatically discover task-specific ``micro-evidence'' that supports faithful reasoning.
We also foresee applications in domains such as mental health assessment, human--computer interaction, and driver monitoring, where trustworthy, human-interpretable reasoning is often more important than raw accuracy alone.

\subsection{Limitations}
\paragraph{Dependence on External AU Detectors.}
While TAG is designed as an AU-grounded reasoning framework, in practice it relies on state-of-the-art automatic AU detectors trained on large-scale human-annotated AU datasets.
Therefore, the grounding signal can be viewed as an approximation to human AU annotations, rather than arbitrary pseudo-labels.
This design choice is motivated by the lack of large-scale datasets that simultaneously provide reliable FER labels, dense AU annotations, and aligned reasoning traces.

Nevertheless, we acknowledge that such detectors still serve only as proxies of true underlying facial muscle activations, and cannot fully capture the physiological processes of facial expression formation.
As a result, TAG does not ground reasoning on direct biomechanical or neuromuscular measurements, but instead on statistically learned AU representations.
Future datasets with high-quality human AU annotations or even physiological sensing signals could further strengthen the biological validity of AU-grounded reasoning and enable more faithful supervision.

\paragraph{Model Choice and Generality.}
For controlled comparison, all our main experiments are conducted with Qwen2.5-VL as the underlying vision--language backbone, which allows us to isolate the effect of TAG-style grounding and training.
However, TAG is not designed specifically for Qwen; rather, it is a general paradigm that can in principle be instantiated with many open-source MLLMs.
Extending TAG to a broader range of backbones (e.g., LLaVA-style, InternVL-style, and larger-scale proprietary models) is an important direction for future work to test the generality and scalability of AU-grounded reasoning.

\paragraph{Efficiency vs.\ Interpretability.}
Compared with traditional CNN or ViT backbone+head FER models, using an MLLM to solve FER inevitably introduces a significant efficiency overhead in terms of computation and latency.
In many real-time scenarios, this overhead may preclude direct deployment if only classification accuracy is considered.
On the other hand, TAG offers rich, step-by-step explanations and explicit AU evidence that black-box models cannot provide, which can be crucial for high-stakes applications such as psychological assessment, clinical diagnosis, and autonomous driving.
We envision future research on model distillation, caching, and lightweight reasoning architectures that retain most of TAG's interpretability benefits while narrowing the efficiency gap with traditional FER models.

\section{Human and LLM Evaluation}
\label{sec:human-llm-eval}

\paragraph{Motivation.}
Beyond standard quantitative metrics such as accuracy and F1, our method explicitly produces reasoning traces that describe how facial action evidence supports the final FER decision.
A natural question is whether these traces are actually aligned with human and LLM preferences, rather than being merely plausible-looking explanations.
To investigate this, we conduct preference-based evaluations that directly compare the reasoning quality of TAG against a strong multimodal baseline.
Concretely, we evaluate TAG-7B after supervised fine-tuning (SFT) on the TAG-310k corpus, without any per-dataset tuning, in order to probe its cross-dataset generalization and robustness.
As the baseline, we adopt the off-the-shelf Qwen2.5-VL-7B model, which is a competitive open-source vision--language model but has not been exposed to the TAG-310k reasoning corpus.
This design isolates the contribution of TAG's task-aligned grounding and large-scale reasoning supervision, while keeping backbone capacity comparable.

\begin{figure}[t]
    \centering
    \includegraphics[width=\linewidth]{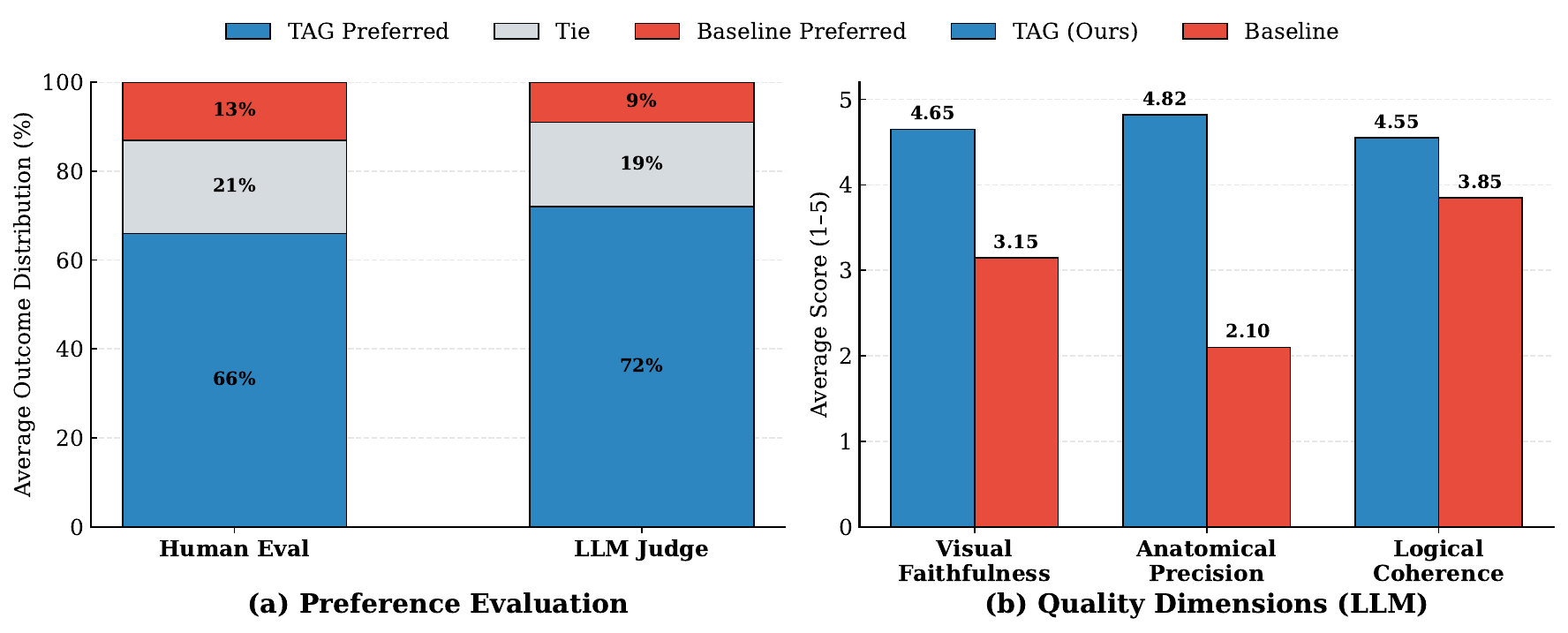}
    \caption{Human and LLM evaluation results.}
    \label{fig:preference-eval}
\end{figure}

\paragraph{Human Preference Evaluation.}
For human evaluation, we first construct paired comparisons between TAG-7B and the baseline.
On each FER evaluation set, we uniformly sample 100 image--question pairs.
For each sampled instance, we generate one reasoning trace and final answer from TAG-7B and one from the baseline, using identical prompts and decoding settings (temperature, max tokens, etc.) per model, and we only use a single sampled response per model to avoid cherry-picking.
This yields one A/B pair of responses for each sampled instance.
We then recruit five human evaluators (two FER experts with domain knowledge in facial action coding, and three non-experts) to conduct anonymized A/B preference judgments.
For each pair, evaluators are shown the input image and the two anonymized reasoning--answer responses, presented in random order and without any model identity information, and are asked to indicate which response they prefer or whether they consider the two responses tied.
Evaluators are instructed to jointly consider the correctness of the final FER prediction, the plausibility and internal consistency of the reasoning trace, and the degree to which the reasoning is grounded in visible facial action cues.
Each evaluator independently labels the full set of A/B pairs; we then aggregate preferences by averaging across the five evaluators.
As summarized in Figure~\ref{fig:preference-eval}, TAG is preferred in 66\% of cases, compared to 13\% for the baseline, with 21\% ties, demonstrating that human judges overwhelmingly favor TAG's generated reasoning and answers over those of the baseline.

\paragraph{LLM-as-a-Judge Evaluation.}
To further obtain a scalable and more fine-grained assessment, we additionally use GPT-5~\cite{openai2025gpt5} (accessed via OpenRouter) as an LLM-as-a-judge.
We adopt a rubric-based evaluation protocol that scores each model's response along three qualitative dimensions: \emph{Visual Faithfulness}, \emph{Anatomical Precision}, and \emph{Logical Coherence}.
Visual Faithfulness measures how well the reasoning refers to observable cues in the given image, avoiding hallucinated content.
Anatomical Precision measures how accurately the reasoning invokes facial musculature and action units in a physiologically meaningful way.
Logical Coherence measures whether the reasoning steps are internally consistent and whether they support the final FER conclusion in a clear, structured manner.
For each sampled image--question pair and its two model responses, we provide GPT-5 with the original prompt, the input image, and both anonymized responses in a fixed, symmetric format, and ask GPT-5 to (i) assign separate 1--5 scores to each response on each dimension (higher is better), and (ii) optionally state an overall preference between the two responses.
We carefully ensure that the prompts, images, and context shown to GPT-5 are identical across models, differing only in the two responses being compared.
To reduce randomness, we repeat the GPT-5 evaluation three times with different random seeds and report averages over the three runs.
Figure~\ref{fig:preference-eval} reports both the preference statistics and the average rubric scores.
GPT-5 prefers TAG in 72\% of cases versus 9\% for the baseline (19\% ties), and TAG achieves higher average scores across all three dimensions (4.65/4.82/4.55) than the baseline (3.15/2.10/3.85).
The particularly strong improvement in Anatomical Precision aligns with TAG's design goal of grounding reasoning in physiologically meaningful facial action evidence, confirming that TAG's reasoning traces are not only preferred but also better aligned with expert knowledge.

\section{Training and Evaluation}
\label{sec:training-eval}

\subsection{Training Configuration}

\paragraph{Supervised Fine-Tuning (SFT).}
We initialize TAG from Qwen2.5-VL-7B and perform supervised fine-tuning on TAG-310K using LLaMA-Factory with LoRA adaptation.
Unless otherwise stated, we freeze the visual encoder and update the projector and language decoder so that both visual features and language reasoning can adapt to AU-grounded supervision.
LoRA is applied to all attention and MLP modules in the language backbone with rank $128$, scaling factor $\alpha = 256$, and LoRA dropout $0.05$.
Following the main text, we use AdamW~\cite{loshchilov2017adamw} with learning rate $2\!\times\!10^{-5}$, cosine decay schedule, weight decay $0.01$, and a warm-up ratio of $3\%$.
We train with a global batch size chosen to fully utilize 4$\times$H200 GPUs (per-GPU batch size $8$ with $8$ gradient accumulation steps), using bfloat16 mixed precision with FP32 master weights.
The maximum context length is $2048$ tokens, covering the full \texttt{<think>} trace, up to three \texttt{<bbox>} blocks, and the final \texttt{<answer>} label.
We clip gradients at a global norm of $1.0$ and train for up to three epochs over TAG-310K, selecting the final checkpoint based on validation accuracy and AU IoU on a held-out split.
This configuration follows standard practice for 7B-scale SFT while being tailored to long, AU-grounded reasoning traces.

\paragraph{Reinforcement Learning (RL).}
On top of the SFT checkpoint, we apply GRPO-based reinforcement learning using the \texttt{verl} framework on the same 4$\times$H200 cluster.
As in the main paper, RL is performed separately for each dataset-specific subset of TAG-310K (RAF-DB, FERPlus, AffectNet), so that the policy can adapt to dataset-specific distribution shifts while preserving the AU-grounded prior from SFT.
For each dataset, we sample $K\!=\!8$ rollouts per prompt and form groups of $G\!=\!8$ trajectories for advantage normalization.
We use AdamW with learning rate $1\!\times\!10^{-6}$, weight decay $0.01$, and gradient clipping at $1.0$.
The GRPO clipping parameter is set to $\epsilon\!=\!0.2$, and the KL penalty coefficient is $0.04$, following standard GRPO configurations~\cite{shao2024grpo}.
We train for $2$ epochs over the corresponding RL subset of TAG-310K for each dataset.
The total reward follows Section~\ref{sec:rl}, $R = R_{\mathrm{AU}} + R_{\mathrm{ans}} + R_{\mathrm{fmt}}$, with rewards normalized within each group before computing policy gradients.
During RL we keep LoRA adapters trainable and continue to freeze the visual encoder, so that optimization focuses on reasoning, formatting, and grounding behavior instead of low-level perception.

\subsection{Evaluation Protocol}

\paragraph{Unified Prompting and CoT Decoding.}
For a fair comparison across all models, we adopt a single unified system prompt that explicitly encourages chain-of-thought reasoning.
The prompt instructs the model to (i) first perform global analysis of the face, (ii) then inspect up to three local regions that are most informative for the expression, and (iii) finally output one categorical expression label.
We use this system prompt Verbatim for TAG, all open-source multimodal baselines, and all closed-source LLMs, and always request ``step-by-step'' reasoning before the final answer.
For TAG and all local open-source VLM baselines, we perform inference using vLLM on a single H200 GPU.
Unless otherwise specified, we set the decoding temperature to $\tau$ (with $\tau\!=\!0$ for TAG and its ablations), top-p $=0.9$, and a maximum generation length of $512$ tokens, which is sufficient to cover the reasoning trace and label.

\paragraph{Closed-Source Models via OpenRouter.}
For closed-source models, we use the OpenRouter API and keep all prompts and decoding parameters as close as possible to the open-source setting.
We evaluate the following models:
\begin{itemize}
    \item \texttt{openai/gpt-5},
    \item \texttt{google/gemini-2.5-flash},
    \item \texttt{google/gemini-2.5-pro}.
\end{itemize}
These models are treated as black-box LLMs without any additional fine-tuning.
Images, questions, system prompts, and user instructions are identical to those used for TAG, so differences in performance and reasoning quality can be attributed to model capabilities rather than prompt design.

\paragraph{Evaluation Splits and Label Space.}
To ensure fair cross-dataset comparison, we focus on the seven basic facial expression categories and remove the \emph{contempt} class from all datasets, so that every model is evaluated under a unified 7-class protocol.
We follow the official evaluation splits of RAF-DB, FERPlus, and AffectNet, and construct test sets by filtering out all samples labeled as \emph{contempt}.
The resulting test set sizes are summarized in Table~\ref{tab:eval-splits}.

\begin{table}[t]
    \centering
    \small
    \begin{tabular}{lcc}
        \toprule
        Dataset & Classes used & \# Test images \\
        \midrule
        RAF-DB & 7 (w/o contempt) & 3{,}068 \\
        FERPlus & 7 (w/o contempt) & 3{,}517 \\
        AffectNet & 7 (w/o contempt) & 3{,}500 \\
        \bottomrule
    \end{tabular}
    \caption{Evaluation splits under the unified 7-class protocol.
    We follow the official evaluation splits of each dataset and exclude the \emph{contempt} category when constructing the test sets.}
    \label{tab:eval-splits}
\end{table}

\subsection{Training Dynamics}
\label{appendix:dynamics}
To better understand how TAG learns AU-grounded reasoning, we analyze the training dynamics during both the SFT and RL stages, tracking FER accuracy, AU IoU, and the individual reward components.

\begin{figure}[t]
    \centering
    \includegraphics[width=\linewidth]{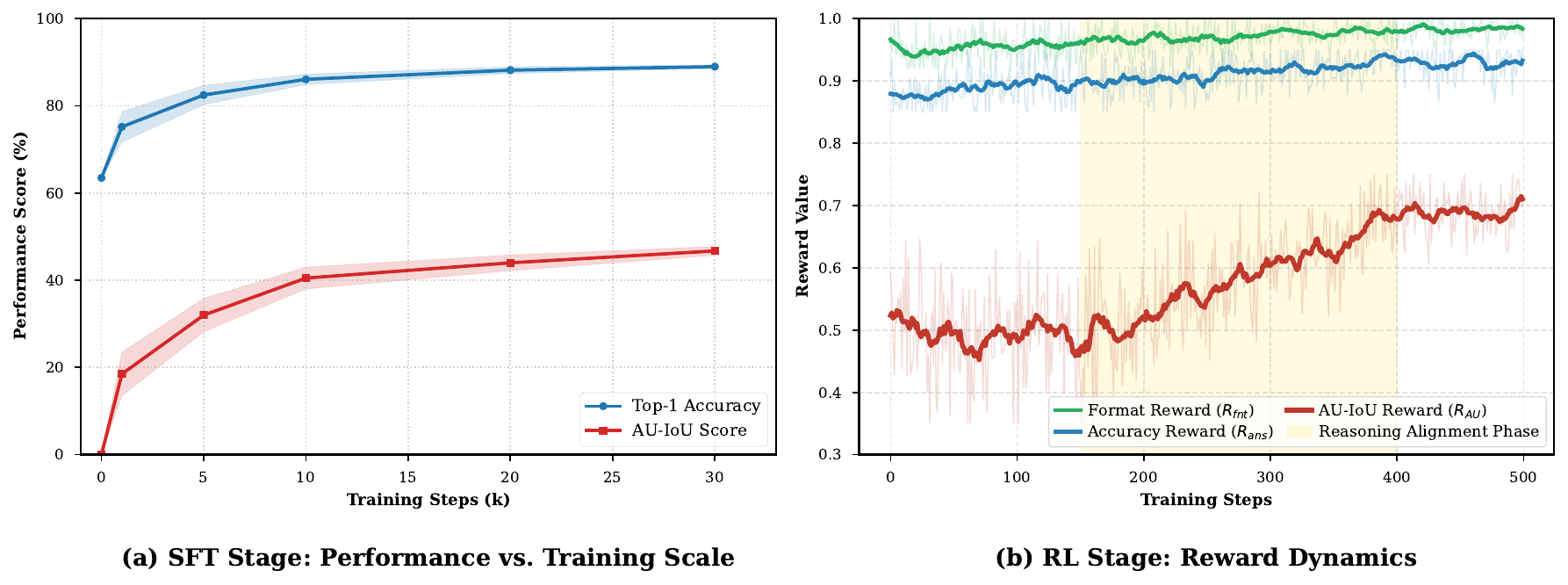}
    \caption{Training dynamics during two stages.}
    \label{fig:training-dynamics}
\end{figure}

\paragraph{SFT Stage.}
Figure~\ref{fig:training-dynamics} shows the training curves during supervised fine-tuning.
As TAG is exposed to more TAG-310K data, both FER accuracy and AU IoU increase rapidly in the early phase and then converge more slowly, reflecting a progression from coarse global understanding to refined AU-level grounding.
In particular, AU IoU continues to improve even after the largest jumps in accuracy, indicating that the model first learns to predict correct labels and then gradually aligns its bounding boxes with the AU detector.
This behavior suggests that SFT establishes a stable AU-grounded reasoning structure rather than memorizing a small set of AU patterns, providing a strong initialization for subsequent RL.

\paragraph{RL Stage.}
During the RL stage, Figure~\ref{fig:training-dynamics} illustrates that FER accuracy quickly saturates on each dataset, while the AU-grounded reward $R_{\mathrm{AU}}$ and the corresponding AU IoU continue to increase.
Once answer accuracy has plateaued, RL still discovers trajectories that yield higher AU overlap and more faithful visual grounding, effectively reshaping the model's internal attention and bounding-box predictions without overfitting label shortcuts.
The fact that $R_{\mathrm{AU}}$ improves after accuracy convergence indicates that TAG is learning more robust, physiologically meaningful features rather than merely pushing for marginal accuracy gains.
Combined with the format reward $R_{\mathrm{fmt}}$, this leads to reasoning traces that are not only correct but also consistently grounded in localized facial evidence, which is essential for interpretable and trustworthy FER.

\section{TAG-310k Dataset}
\label{sec:appendix-a}
\subsection{Overview}
TAG-310K is a large-scale, reasoning-augmented facial expression dataset consisting of 318,269 samples collected from three widely used FER benchmarks: AffectNet, FERPlus, and RAF-DB.
Each sample is associated with (i) a facial image, (ii) a categorical expression label, (iii) a set of Action Units (AUs) with spatial bounding boxes, and (iv) a structured, multi-step reasoning trace produced by a vision--language model.
The dataset is designed to support research on interpretable facial expression recognition, where decisions are grounded in localized facial muscle evidence rather than implicit global features.

\subsection{Generation Details}
\label{sec:appendix-generation}
For TAG-310k construction, we employ a distributed generation pipeline using 4$\times$H200 GPUs, with each GPU serving one Qwen2.5-VL-32B model instance.
We use 512 concurrent requests, processing them serially through three stages: VL-Quality filtering, API access for reasoning trace generation, and retry handling.
Each GPU handles 128 concurrent requests, ensuring efficient resource utilization while maintaining generation quality.
The detailed pipeline follows the multi-stage process described in Section~\ref{sec:data}, including quality checks, top-$K$ AU-based CoT generation, format validation, and retry mechanisms for incorrect predictions.
The overall procedure is further summarized in Algorithm~\ref{alg:tag310k_generation}.

\begin{algorithm}[h]
    \caption{Distributed multi-stage generation pipeline for TAG-310k.}
    \label{alg:tag310k_generation}
    \DontPrintSemicolon
    \SetKwInOut{Input}{Input}
    \SetKwInOut{Output}{Output}
    
    \Input{Training split images with labels $\{(x_i, y_i)\}$; AU/landmark detectors; VLM/VL-quality checker;
    Qwen2.5-VL-32B API for reasoning trace generation.}
    \Output{Accepted samples $\mathcal{D}$ with AU-grounded traces in \texttt{<think>}/\texttt{<bbox>}/\texttt{<answer>} format.}
    
    \BlankLine
    \textbf{System setup:} allocate $4\times$H200 GPUs; launch one Qwen2.5-VL-32B instance per GPU.\;
    Set total concurrency $C{=}512$ with per-GPU concurrency $C_g{=}128$.\;
    Initialize a global request queue $\mathcal{Q}$ with all candidate samples.\;
    Initialize accepted set $\mathcal{D}\leftarrow \emptyset$.\;
    
    \BlankLine
    \While(\tcp*[f]{Workers run in parallel on 4 GPUs}){$\mathcal{Q}$ not empty}{
      Dequeue a mini-batch $\mathcal{B}$ of up to $C_g$ samples for this GPU.\;
    
      \BlankLine
      \tcp{Stage I: VL-quality filtering}
      $\mathcal{B}_{\text{ok}}\leftarrow \emptyset$.\;
      \ForEach{$(x,y)\in \mathcal{B}$}{
        \If{$\operatorname{VLQuality}(x) = \texttt{pass}$}{
          $\mathcal{B}_{\text{ok}}\leftarrow \mathcal{B}_{\text{ok}}\cup\{(x,y)\}$\;
        }
      }
    
      \BlankLine
      \tcp{Stage II: reasoning trace generation (API call)}
      \ForEach{$(x,y)\in \mathcal{B}_{\text{ok}}$}{
        Obtain top-$K$ activated AUs and boxes $\mathcal{A}_K(x)$ from external detectors.\;
        $r \leftarrow \operatorname{GenerateTrace}(x, \mathcal{A}_K(x))$ \tcp*[f]{Qwen2.5-VL-32B}\;
    
        \BlankLine
        \tcp{Stage III: format validation and retry handling}
        \If{$\neg \operatorname{ValidFormat}(r)$}{
          \Repeat{$\operatorname{ValidFormat}(r)$}{
            $r \leftarrow \operatorname{RetryFormat}(x, \mathcal{A}_K(x))$\;
          }
        }
        \If{$\operatorname{PredLabel}(r) \neq y$}{
          Initialize candidate label set $\mathcal{Y}\leftarrow \mathcal{Y}_{\text{all}}$.\;
          \While{$\operatorname{PredLabel}(r)\neq y$ \textbf{and} $\mathcal{Y}\setminus\{\operatorname{PredLabel}(r)\}\neq \emptyset$}{
            $\mathcal{Y}\leftarrow \mathcal{Y}\setminus\{\operatorname{PredLabel}(r)\}$ \tcp*[f]{Iterative label elimination}\;
            $r \leftarrow \operatorname{ReGenerateWithCandidates}(x, \mathcal{A}_K(x), \mathcal{Y})$\;
            \If{$\neg \operatorname{ValidFormat}(r)$}{
              \Repeat{$\operatorname{ValidFormat}(r)$}{
                $r \leftarrow \operatorname{RetryFormat}(x, \mathcal{A}_K(x))$\;
              }
            }
          }
        }
    
        \BlankLine
        \If{$\operatorname{ValidFormat}(r)$ \textbf{and} $\operatorname{PredLabel}(r)=y$}{
          $\mathcal{D}\leftarrow \mathcal{D}\cup\{(x,y,r,\mathcal{A}_K(x))\}$\;
        }
      }
    }
    \Return{$\mathcal{D}$}\;
    \end{algorithm}

\subsection{Dataset Sources and Class Distribution}

\begin{figure*}[p]
    \centering

    % ---- Row 1: A1 full width ----
    \begin{subfigure}[t]{0.98\textwidth}
        \centering
        \includegraphics[width=\linewidth]{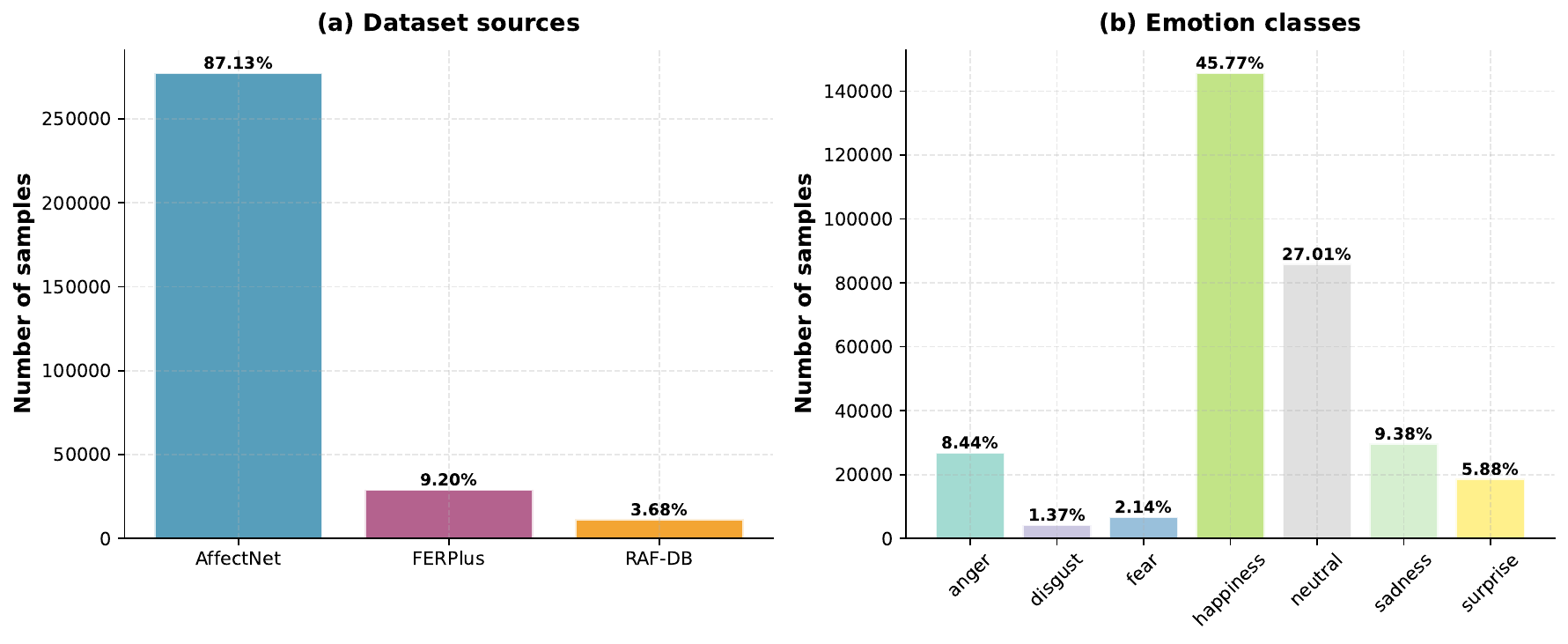}
        \caption{Dataset sources and emotion classes.}
        \label{fig:appendix-a8}
    \end{subfigure}

    \vspace{0.8em}

    % ---- Row 2: A2 + A3 ----
    \begin{subfigure}[t]{0.49\textwidth}
        \centering
        \includegraphics[width=\linewidth]{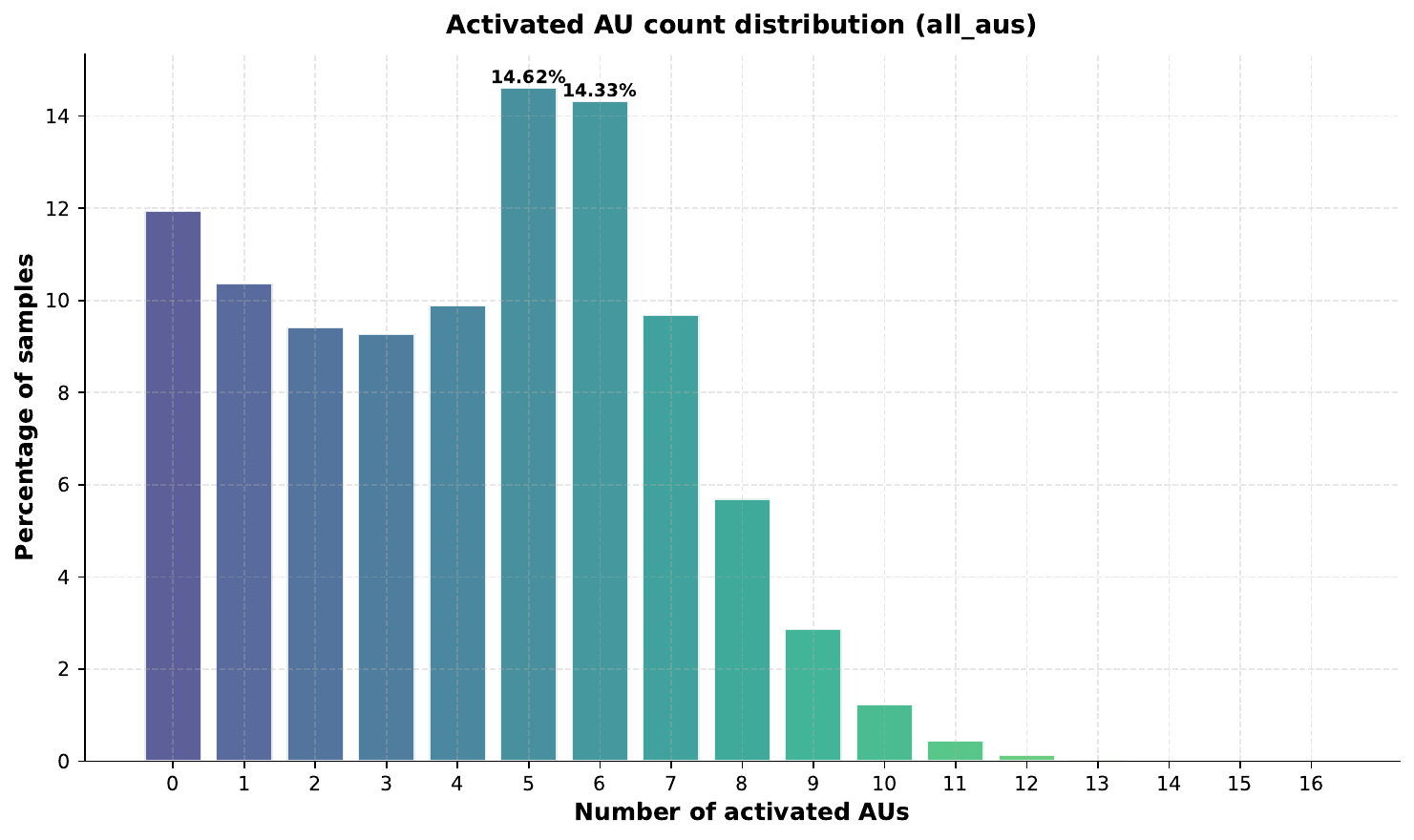}
        \caption{Distribution of activated AU counts per sample.}
        \label{fig:appendix-a9}
    \end{subfigure}
    \hfill
    \begin{subfigure}[t]{0.49\textwidth}
        \centering
        \includegraphics[width=\linewidth]{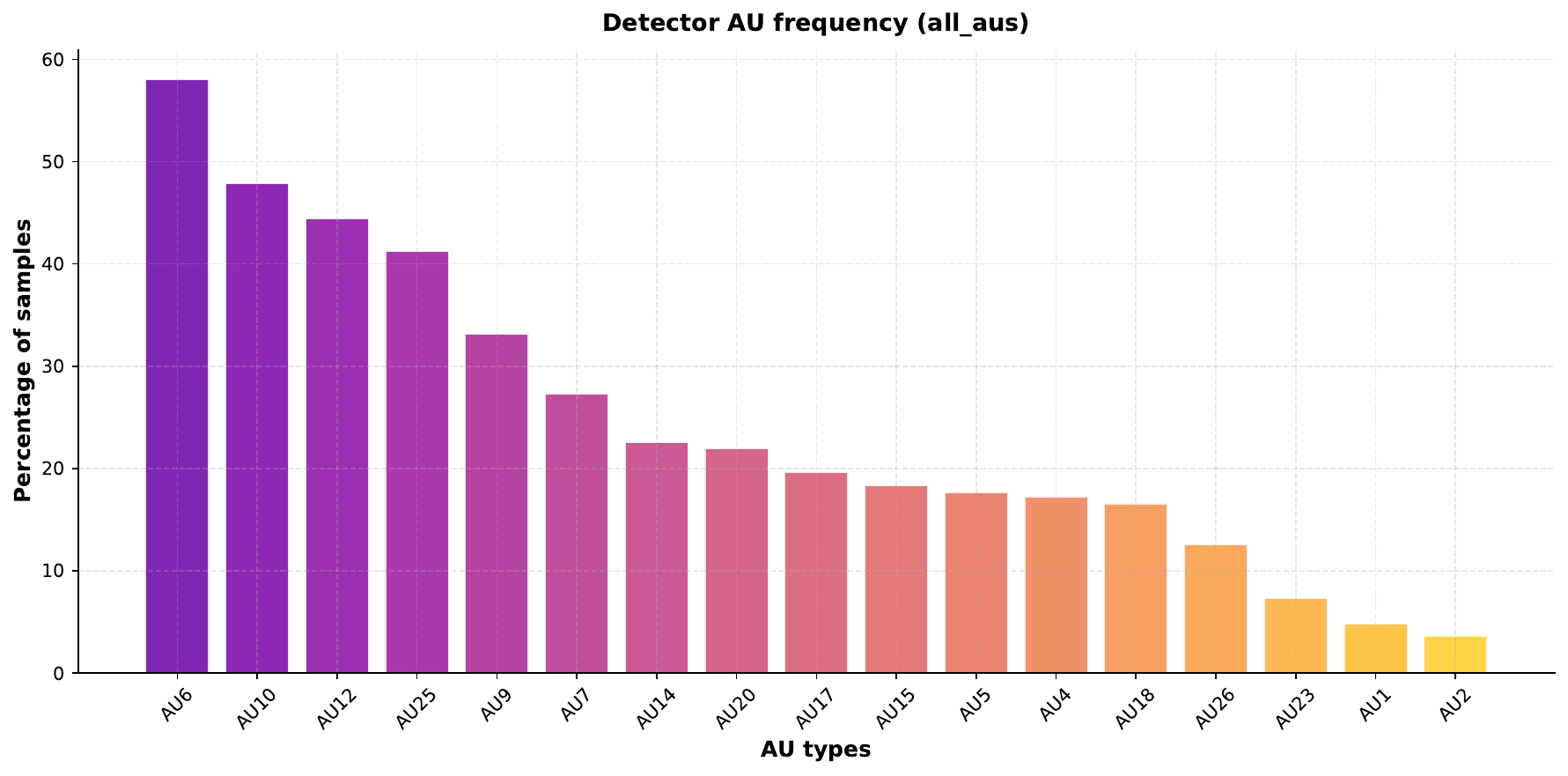}
        \caption{AU type frequency across all samples.}
        \label{fig:appendix-a10}
    \end{subfigure}

    \vspace{0.8em}

    % ---- Row 3: A4 + A5 ----
    \begin{subfigure}[t]{0.49\textwidth}
        \centering
        \includegraphics[width=\linewidth]{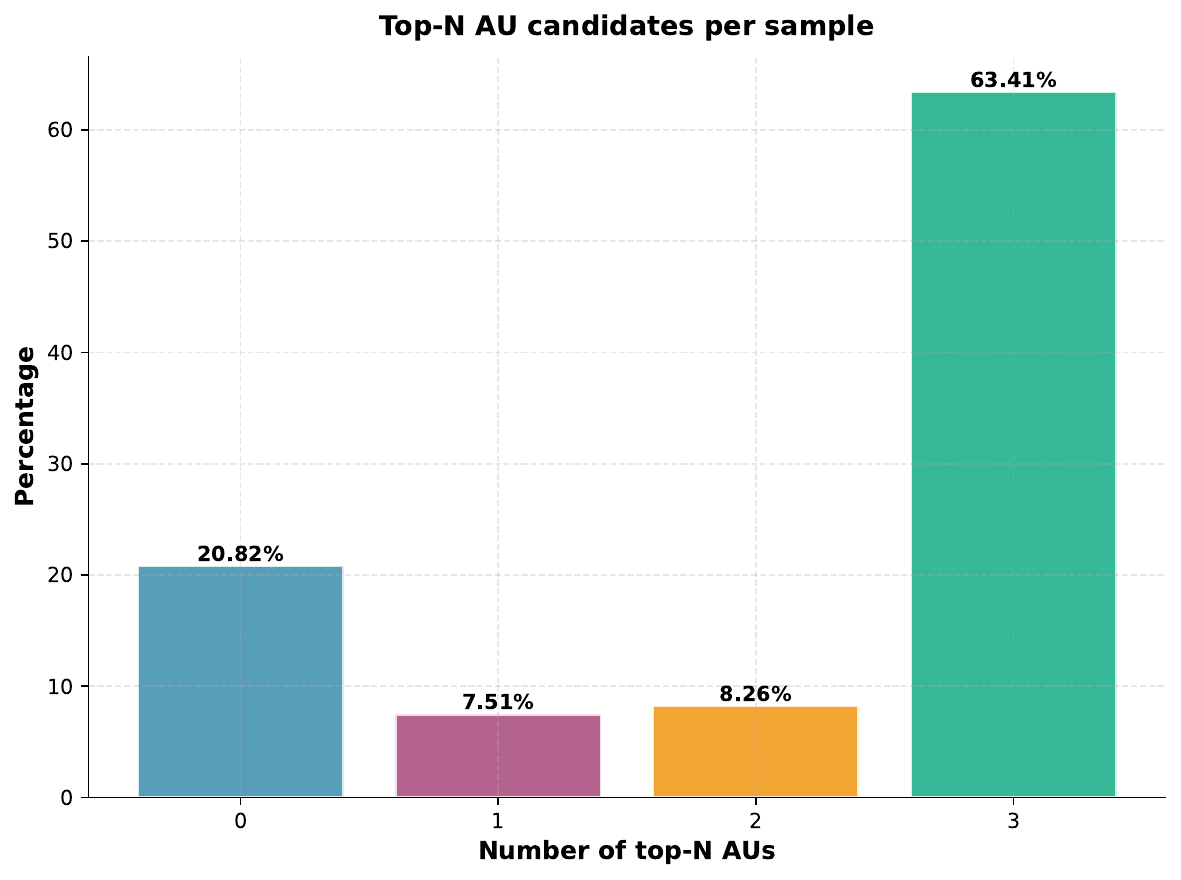}
        \caption{Distribution of top-$N$ candidate AUs per sample.}
        \label{fig:appendix-a11}
    \end{subfigure}
    \hfill
    \begin{subfigure}[t]{0.49\textwidth}
        \centering
        \includegraphics[width=\linewidth]{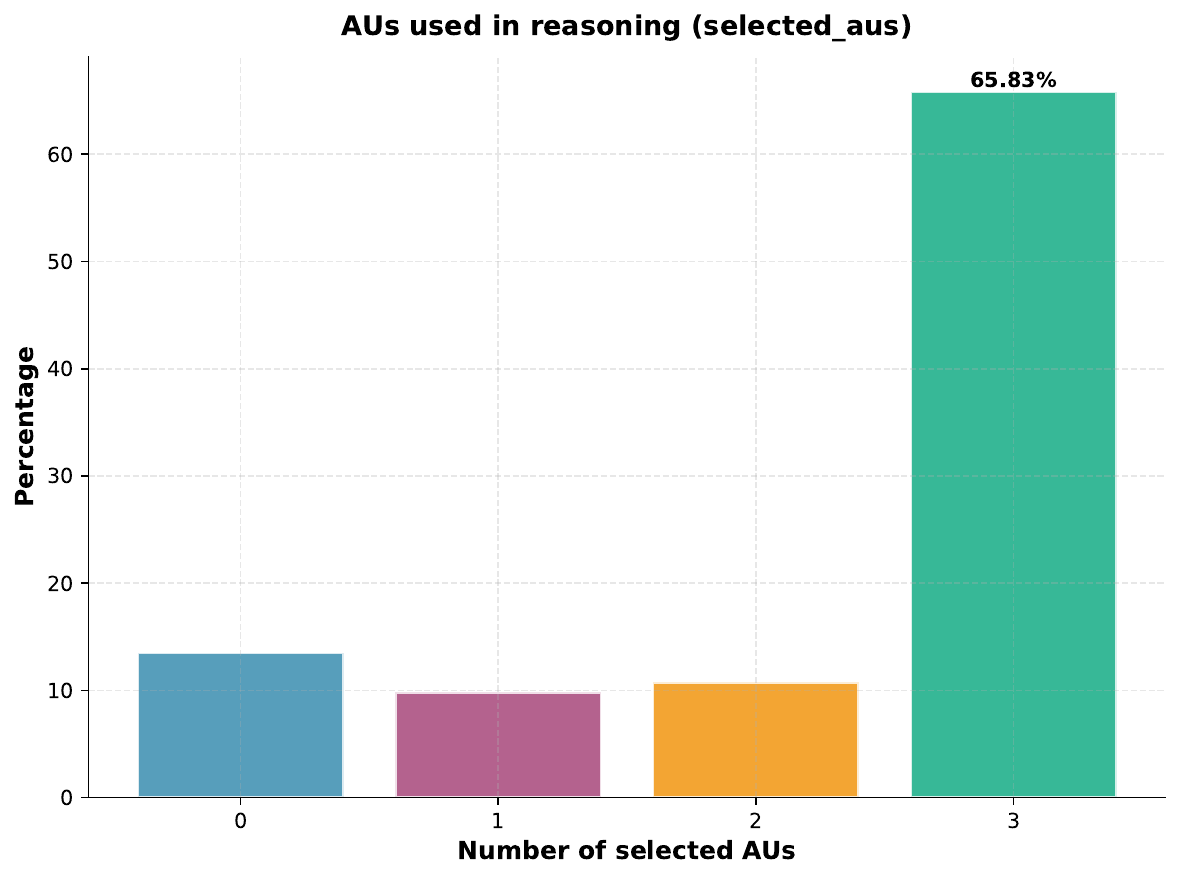}
        \caption{Distribution of selected AUs used in reasoning.}
        \label{fig:appendix-a12}
    \end{subfigure}

    \caption{TAG-310K appendix statistics: dataset composition and AU-related distributions.}
    \label{fig:appendix-a_all}
\end{figure*}

\paragraph{Dataset Composition.}
TAG-310K integrates samples from three datasets: AffectNet (87.13\%), FERPlus (9.20\%), and RAF-DB (3.68\%).
The dataset composition and emotion class distribution are shown in Fig.~\ref{fig:appendix-a8}.

\paragraph{Expression Class Distribution.}
The distribution over the seven basic expression categories reflects the natural imbalance present in real-world facial expression data, with happiness and neutral being the most frequent classes.
We intentionally preserve this imbalance to avoid introducing artificial priors that may distort real-world deployment behavior.

\paragraph{Detector-Level AU Statistics.}
For each image, an AU detector produces a dense set of activated AUs (all\_aus), representing all facial muscle movements detected above a minimal confidence threshold.
As shown in Fig.~\ref{fig:appendix-a9}, most samples activate multiple AUs, with an average of approximately 5--6 AUs per image, indicating rich but maybe noisy facial muscle activations.
The frequency of each AU type across all samples is shown in Fig.~\ref{fig:appendix-a10}.

\paragraph{Candidate AU Pool.}
To reduce reasoning complexity while preserving salient evidence, we construct a candidate AU pool by selecting the top-$N$ AUs based on detector confidence, with $N=3$ chosen as a conservative trade-off between evidence coverage and search space size.
This step does not enforce a fixed reasoning structure, but merely constrains the initial search space presented to the model.

As shown in Fig.~\ref{fig:appendix-a11}, more than 63\% of samples contain exactly three candidate AUs, while approximately 21\% contain no confident candidates under the top-$N$ selection rule.

\paragraph{AU Selection in Final Reasoning.}
\label{sec:appendix-a12}
During reasoning, the model may select zero to multiple AUs from the candidate pool to construct its final explanation.
As illustrated in Fig.~\ref{fig:appendix-a12}, the model predominantly relies on no more than three AUs, with over 65\% of samples using exactly three AUs, demonstrating a strong preference for concise and sparse evidence.
This sparsity emerges naturally from the reasoning process and is not enforced by hard constraints, suggesting that the reasoning paths prioritize minimal yet sufficient evidence.

\subsection{Reasoning Length and Token Statistics}
\begin{figure}[!t]
    \centering
    \includegraphics[width=0.5\textwidth]{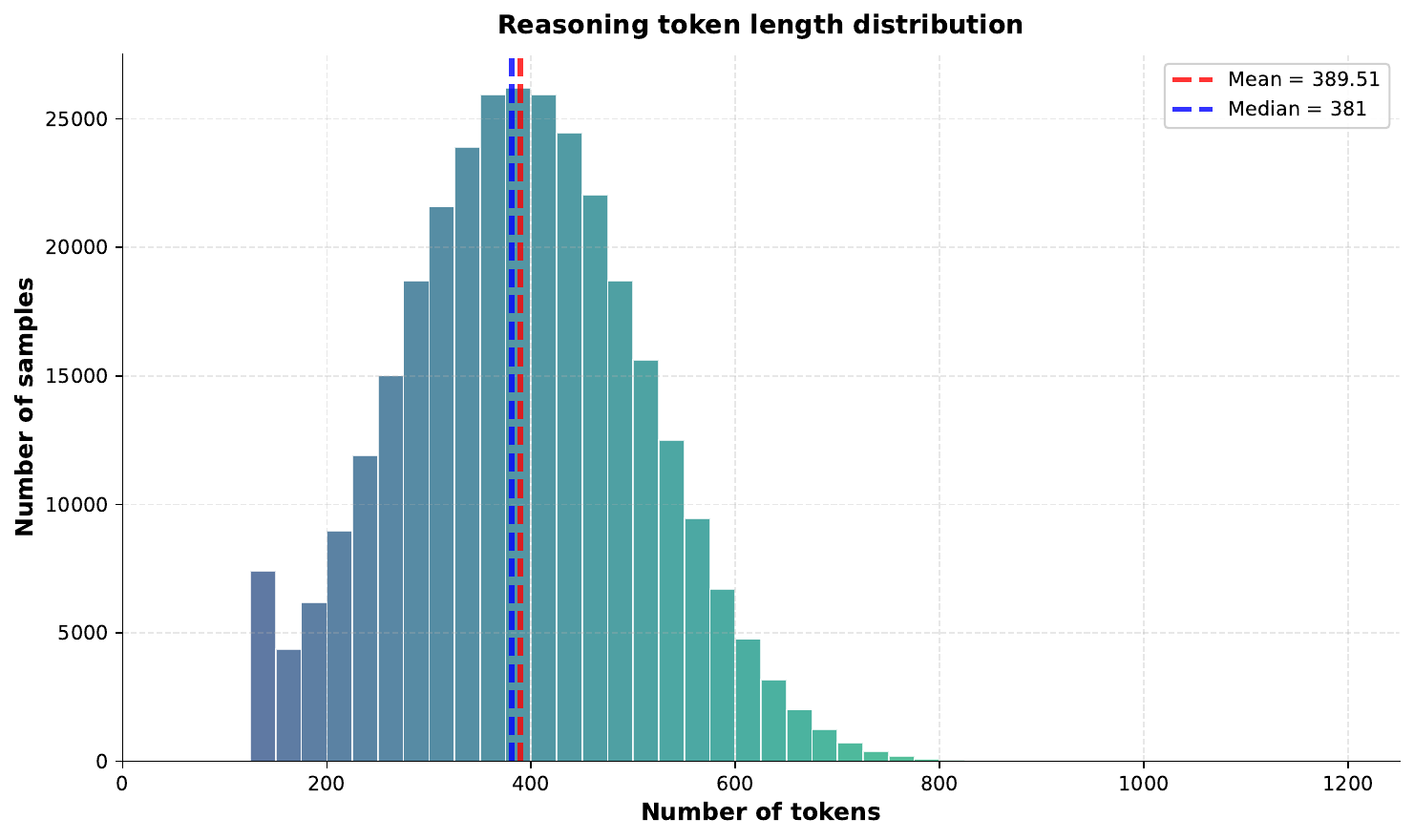}
    \caption{Distribution of reasoning trace lengths (in tokens).}
    \label{fig:appendix-a13}
\end{figure}

Each TAG-310K sample contains a structured multi-step reasoning trace.
The average reasoning length is 389 tokens, with a maximum of 1221 tokens, indicating substantial compositional reasoning rather than short templated responses (see Fig.~\ref{fig:appendix-a13}).
This variability reflects differences in expression complexity and AU evidence availability across samples.

\section{Inference-Time Grounding Analysis}
\label{sec:appendix-c}

During inference, TAG does not have access to any external AU detector outputs.
All bounding boxes are generated autoregressively by the model itself.
In this section, we analyze how the learned Thinking with AU Grounding behavior manifests at inference time, focusing on consistency with external detectors when they are reliable, and emergent grounding beyond detector coverage.

\subsection{Consistency with External AU Detectors}
\label{sec:appendix-c1}
\begin{figure*}[t]
    \centering

    % RAF-DB row
    \begin{subfigure}[t]{0.24\textwidth}
        \centering
        \includegraphics[width=\linewidth]{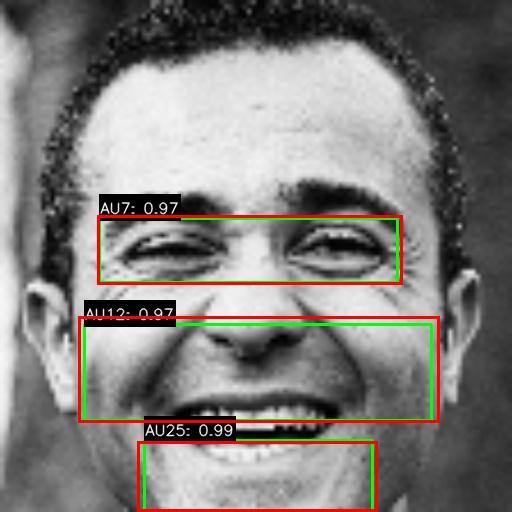}
        \caption{}
        \label{fig:raf-a1}
    \end{subfigure}
    \hfill
    \begin{subfigure}[t]{0.24\textwidth}
        \centering
        \includegraphics[width=\linewidth]{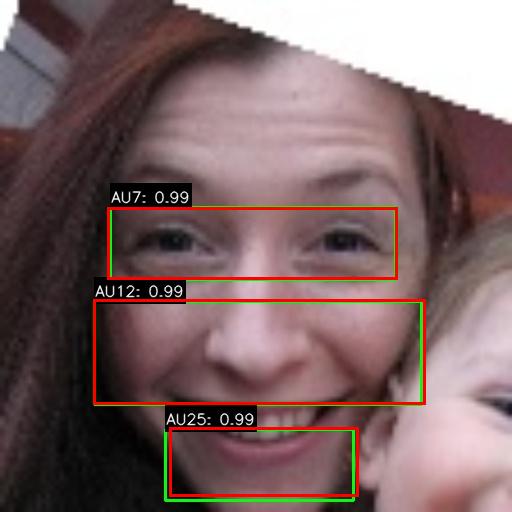}
        \caption{}
        \label{fig:raf-a2}
    \end{subfigure}
    \hfill
    \begin{subfigure}[t]{0.24\textwidth}
        \centering
        \includegraphics[width=\linewidth]{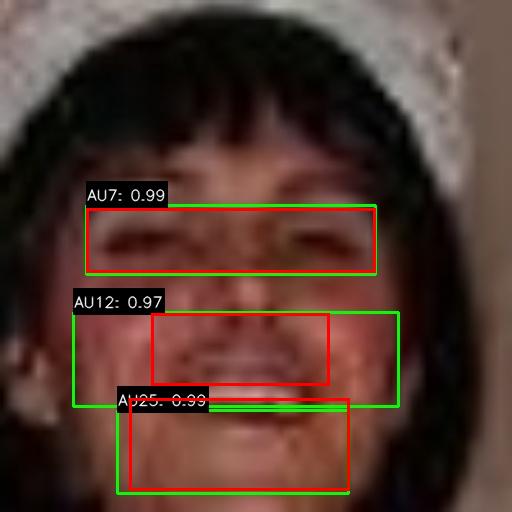}
        \caption{}
        \label{fig:raf-b1}
    \end{subfigure}
    \hfill
    \begin{subfigure}[t]{0.24\textwidth}
        \centering
        \includegraphics[width=\linewidth]{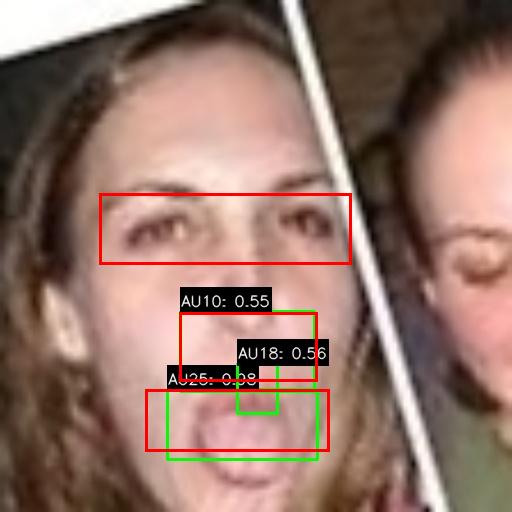}
        \caption{}
        \label{fig:raf-b2}
    \end{subfigure}

    \vspace{0.6em}

    % FERPlus row
    \begin{subfigure}[t]{0.24\textwidth}
        \centering
        \includegraphics[width=\linewidth]{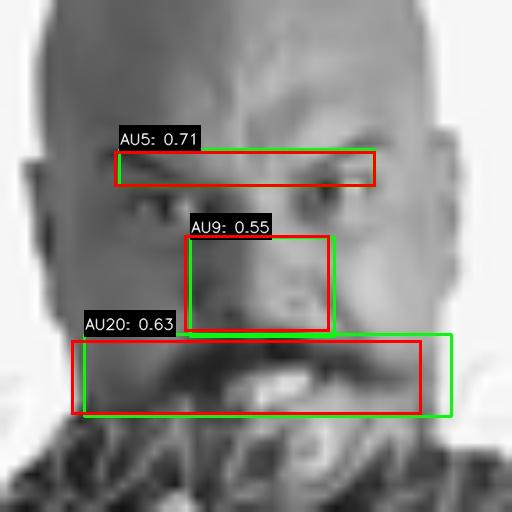}
        \caption{}
        \label{fig:fer-a1}
    \end{subfigure}
    \hfill
    \begin{subfigure}[t]{0.24\textwidth}
        \centering
        \includegraphics[width=\linewidth]{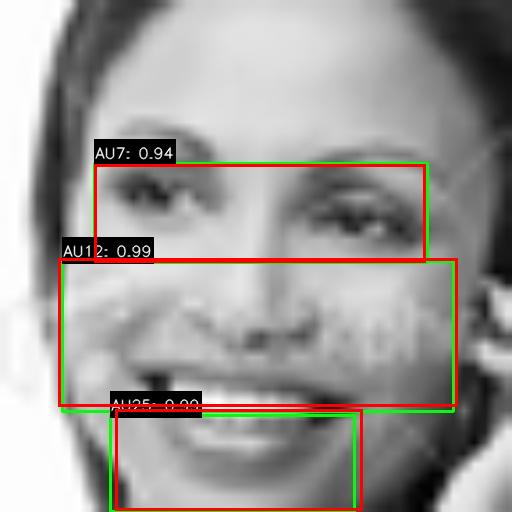}
        \caption{}
        \label{fig:fer-a2}
    \end{subfigure}
    \hfill
    \begin{subfigure}[t]{0.24\textwidth}
        \centering
        \includegraphics[width=\linewidth]{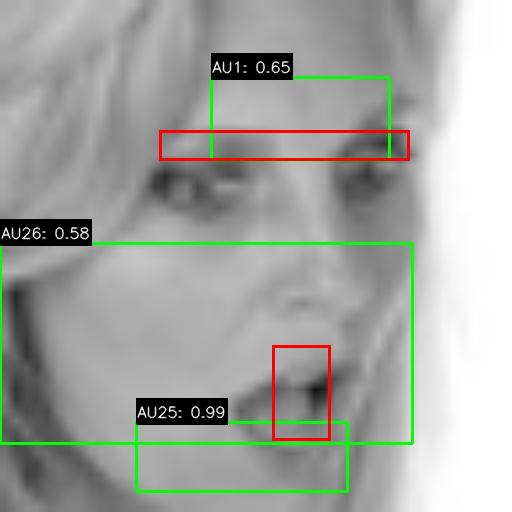}
        \caption{}
        \label{fig:fer-b1}
    \end{subfigure}
    \hfill
    \begin{subfigure}[t]{0.24\textwidth}
        \centering
        \includegraphics[width=\linewidth]{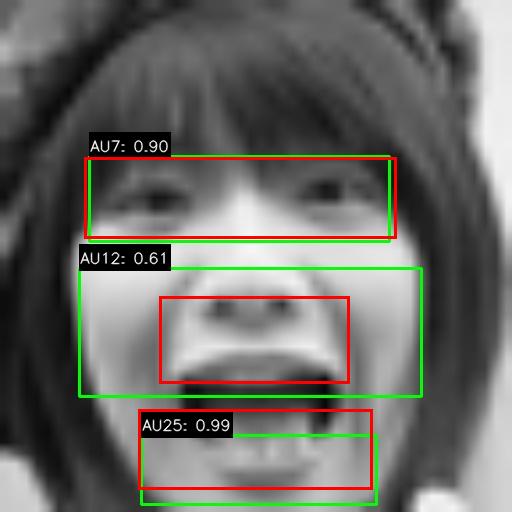}
        \caption{}
        \label{fig:fer-b2}
    \end{subfigure}

    \vspace{0.6em}

    % AffectNet row
    \begin{subfigure}[t]{0.24\textwidth}
        \centering
        \includegraphics[width=\linewidth]{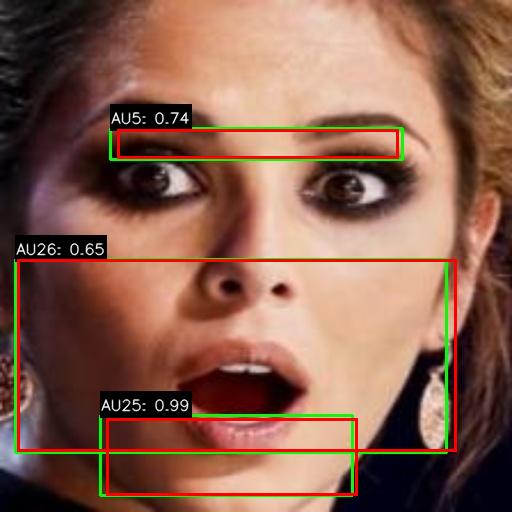}
        \caption{}
        \label{fig:affectnet-a1}
    \end{subfigure}
    \hfill
    \begin{subfigure}[t]{0.24\textwidth}
        \centering
        \includegraphics[width=\linewidth]{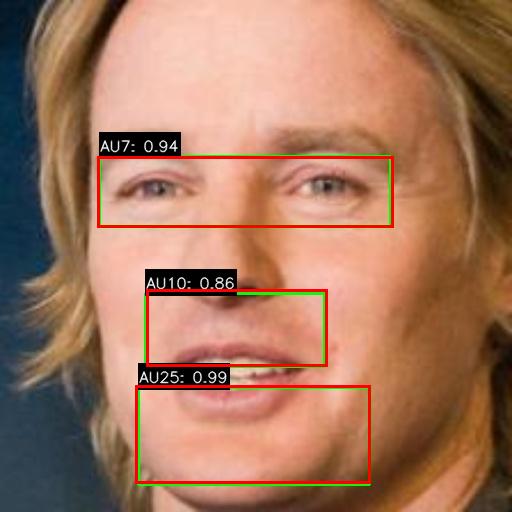}
        \caption{}
        \label{fig:affectnet-a2}
    \end{subfigure}
    \hfill
    \begin{subfigure}[t]{0.24\textwidth}
        \centering
        \includegraphics[width=\linewidth]{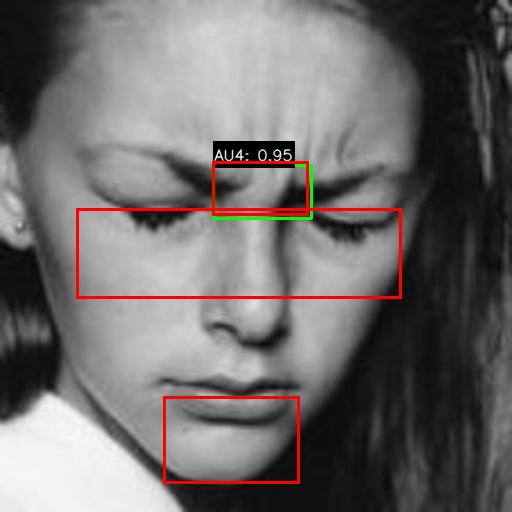}
        \caption{}
        \label{fig:affectnet-b1}
    \end{subfigure}
    \hfill
    \begin{subfigure}[t]{0.24\textwidth}
        \centering
        \includegraphics[width=\linewidth]{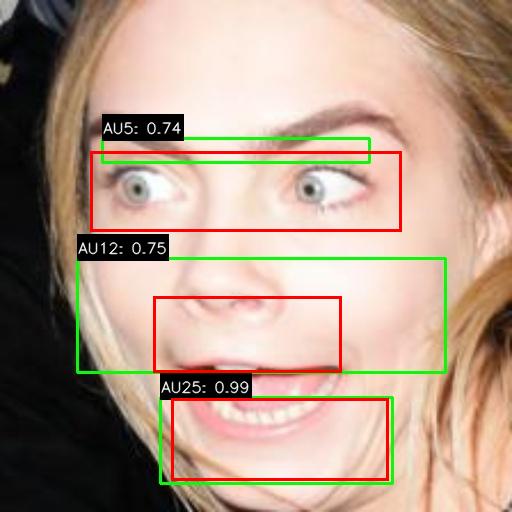}
        \caption{}
        \label{fig:affectnet-b2}
    \end{subfigure}

    \caption{Inference-time grounding analysis across RAF-DB (top row), FERPlus (middle row), and AffectNet (bottom row). Left two columns (a) and (b) show cases where TAG's predictions align with external AU detectors, demonstrating consistency. Right two columns (c) and (d) show cases where TAG attends to regions beyond external detector coverage, demonstrating emergent grounding capability. Green boxes indicate external AU detector outputs with confidence scores, while red boxes represent TAG's predicted regions.}
    \label{fig:inference-grounding-all}
\end{figure*}

\paragraph{Overview.}
When external AU detectors reliably identify salient facial muscle activations, TAG often predicts bounding boxes that spatially align with these regions, despite having no access to detector outputs at inference time.
This indicates that TAG internalizes physiologically meaningful facial patterns through AU-grounded supervision, rather than relying on detector signals during prediction.
As shown in Figure~\ref{fig:inference-grounding-all}(a) and (b), (e) and (f), and (i) and (j), TAG's predicted regions consistently align with external AU detector outputs, demonstrating that the model learns to attend to the same physiologically meaningful facial regions identified by external detectors.

\subsection{Beyond External Detectors}
\label{sec:appendix-c2}

In more challenging cases, external AU detectors fail to activate or miss subtle but discriminative facial regions.
Notably, TAG can still attend to visually and anatomically plausible regions that support correct predictions, even when these regions are absent from detector outputs.
This behavior demonstrates that TAG does not merely replicate external detectors, but instead learns a generalized grounding capability that extrapolates beyond detector limitations.

\paragraph{More Precise Grounding.}
TAG can produce more precise grounding than external detectors in certain cases.
For instance, in Figure~\ref{fig:inference-grounding-all}(c), (g) and (h), and (l), TAG identifies finer-grained facial regions that are more discriminative for expression recognition, even when external detectors provide coarser or less accurate localizations.

\paragraph{Human-Aligned Grounding.}
TAG's predictions can align more closely with human-annotated ground truth than external detectors.
As shown in Figure~\ref{fig:inference-grounding-all}(d) and (g), TAG attends to regions that are anatomically meaningful and consistent with human perception of facial expression cues, demonstrating improved alignment with human understanding of expression-related facial regions.

\paragraph{Grounding under Detector Failure.}
When external AU detectors fail to activate or produce unreliable outputs, TAG can still provide meaningful grounding.
For example, in cases where detectors miss subtle but important facial cues, TAG successfully identifies relevant regions that support correct predictions, as illustrated in Figure~\ref{fig:inference-grounding-all}(k).
This capability is particularly valuable in real-world scenarios where detector reliability may vary.

\clearpage
\section{Prompts}
\label{sec:appendix-b}

This appendix details the prompt protocols used in TAG for both dataset construction and model evaluation.
Rather than treating prompts as task-specific heuristics, we view them as interface constraints that regulate reasoning structure, evidence usage, and output format.
All reported performance gains are obtained under fixed prompts during evaluation, ensuring that improvements arise from learned grounding behavior rather than prompt variation.

\begin{promptbox}{VL-Quality Filtering Prompt (Image Quality Check)}
    %\footnotesize
    \begin{Verbatim}[breaklines,breakanywhere]
  # Role & Objective
  You are a professional image quality assessment expert. Your task is to evaluate whether a face image `<image>` is suitable for training a facial expression recognition model.
  
  # Quality Criteria
  An image is suitable for training if it meets ALL of the following criteria:
  1. **Face Visibility**: The face is clearly visible and not heavily occluded
  2. **Image Quality**: The image is not blurry, too dark, too bright, or severely corrupted
  3. **Face Completeness**: The face is not cropped or cut off in a way that loses important facial features
  4. **Expression Clarity**: The facial expression is reasonably clear and not ambiguous
  5. **No Severe Artifacts**: The image does not have severe compression artifacts, watermarks, or other issues that would interfere with training
  
  # Output Format (Strict JSON Only)
  You MUST output ONLY a valid JSON object:
  
  {
    "suitable_for_training": true/false,
    "reason": "Brief explanation of why the image is or is not suitable for training"
  }
  
  # Constraints
  1. Be strict but fair in your assessment
  2. Only mark as suitable if the image clearly meets all criteria
  3. If in doubt, err on the side of caution (mark as not suitable)
  4. The reason should be concise but informative
    \end{Verbatim}
  \end{promptbox}

  \newpage

\begin{promptbox}{TAG AU-Grounded Generation Prompt}
  %\footnotesize
  \begin{Verbatim}[breaklines,breakanywhere]
# Role & Objective
You are a professional facial expression analysis expert. Your task is to generate a complete decision-making process containing **reasoning, tool markers, and final prediction** based on a face image `<image>` and internally detected Action Unit (AU) data.

# Internal AU Data (Invisible to the User)
The following AU detections represent hidden prior knowledge.  
They MUST ONLY be used in two ways:
1) as region markers for zoom inspection, and  
2) as semantic labels that describe the visually observed muscle movement.

### ABSOLUTE RESTRICTION
You MUST NOT:
- invent new AUs,
- reference any AU not listed in AU_LIST,
- modify AU numbers,
- modify, reorder, or alter bounding box coordinates,
- create additional bounding boxes.

You may ONLY use AUx and its EXACT bounding box values as provided in AU_LIST.

AU Data List:
{AU_LIST}

# Expression Categories
[anger, disgust, fear, happiness, neutral, sadness, surprise]

# Cognitive Principle
Your reasoning must mimic a human expert:
1. First perform a **global visual analysis** (no AU references, no tool use).
2. Decide which regions to zoom into **based only on your visual impression**.
3. After zooming, describe what you see and optionally name the AU corresponding to that movement.

AU values MUST NEVER be used as evidence or decision-making signals.

---

# Required Structure for the "CoT" Field

## 1. Global Analysis (Start)
Describe:
- The global facial appearance,
- Initial hypotheses,
- Which regions seem visually important and **why** (based only on intuition).

You MUST NOT reference any AU in this section.

---

## 2. Verification Loop (Iterative)

For each region you choose to inspect, follow this **strict ordered sequence**:

### (1) Pre-Tool Reasoning
Explain why this region looks visually relevant.  
This must come ONLY from your visual intuition.

### (2) Tool Trigger (AU Tag)
Insert the AU tag for the region:

    <AUx>

The AU number MUST come from AU_LIST.  
No new AUs may be introduced.

### (3) Tool Arguments (Bounding Box)
Immediately after the AU tag, insert the bounding box EXACTLY as given in AU_LIST:

    [x1, y1, x2, y2]

### STRICT FORMAT REQUIREMENT
The full tool marker **must appear exactly** as:

    <AUx> [x1, y1, x2, y2]

No characters may be added, removed, or reordered.  
Coordinates must match AU_LIST exactly.

### (4) Post-Tool Observation
After zooming:
- Describe what you visually observe,
- Then optionally state that this movement corresponds to AUx.

You MUST NOT:
- use AU values or confidence,
- claim numerical AU evidence,
- introduce any AU not in AU_LIST.

Repeat this loop for every inspected region.

---

## 3. Final Conclusion (End)
Summarize all observations and explain how they lead to the final expression category.

Then output the final predicted category.

---

# Output Format (Strict JSON Only)
You MUST output ONLY a valid JSON object:

{
  "CoT": "Global analysis... <AUx> [x1, y1, x2, y2] local observation... Final conclusion...",
  "Answer": "Expression_Category"
}

---

# Hard Constraints Summary
1. Do NOT introduce new AUs or new bounding boxes.  
2. You may ONLY use AUx tags and bbox values from AU_LIST exactly as given.  
3. The `<AUx> [bbox]` format is **strict and must never be altered**.  
4. Global analysis MUST NOT mention AUs.  
5. AU MUST NOT influence which regions you choose to inspect.  
6. AU is used only for:
   - zooming marker,
   - naming muscle movement AFTER visual observation.
7. `Answer` must be exactly one category from the provided list.
\end{Verbatim}
\end{promptbox}

\newpage

\begin{promptbox}{TAG No-AU Generation Prompt}
    %\footnotesize
    \begin{Verbatim}[breaklines,breakanywhere]
  # Role & Objective
  You are a professional facial expression analysis expert. Your task is to analyze a face image `<image>` and determine the emotional expression.
  
  # Expression Categories
  [anger, disgust, fear, happiness, neutral, sadness, surprise]
  
  # Cognitive Principle
  Your reasoning must mimic a human expert:
  1. First perform a **global visual analysis** of the facial features.
  2. Observe key regions such as eyes, mouth, eyebrows, and overall facial tension.
  3. Based on your observations, determine the emotional expression.
  
  ---
  
  # Required Structure for the "CoT" Field
  
  ## 1. Global Analysis (Start)
  Describe:
  - The global facial appearance,
  - Initial observations about key facial features,
  - Which regions seem important for emotion recognition.
  
  ---
  
  ## 2. Detailed Observation
  Describe what you observe in the key facial regions:
  - Eye region (eyebrows, eyelids, gaze direction)
  - Mouth region (lip position, curvature, openness)
  - Overall facial tension and muscle activity
  
  ---
  
  ## 3. Final Conclusion (End)
  Summarize all observations and explain how they lead to the final expression category.
  
  Then output the final predicted category.
  
  ---
  
  # Output Format (Strict JSON Only)
  You MUST output ONLY a valid JSON object:
  
  {
    "CoT": "Global analysis... Detailed observations... Final conclusion...",
    "Answer": "Expression_Category"
  }
  
  ---
  
  # Constraints Summary
  1. Base your analysis only on visual observations from the image.
  2. Do not invent or reference Action Units (AUs) unless you have specific knowledge.
  3. `Answer` must be exactly one category from the provided list.
    \end{Verbatim}
  \end{promptbox}

  \newpage
\begin{promptbox}{LLM-as-a-Judge Prompt for Human and LLM Evaluation}
  %\footnotesize
  \begin{Verbatim}[breaklines,breakanywhere]
You are an expert judge for facial expression reasoning quality. Your task is to evaluate and compare two anonymized model responses (Response A and Response B) to the same facial expression recognition (FER) problem, given the input image and the original user query.

Your evaluation must follow a strict rubric along three dimensions:

1. Visual Faithfulness (1–5):
   - How well does the reasoning refer to actually visible facial cues in the image (eyes, eyebrows, mouth, wrinkles, overall tension, etc.)?
   - Does it avoid hallucinating non-existent details (e.g., describing occluded regions, imaginary accessories, or lighting conditions that are not present)?

2. Anatomical Precision (1–5):
   - How accurately does the reasoning describe facial musculature and action-related behavior (e.g., brow raising, lip corner pulling, eyelid tightening), even if it does not explicitly name AUs?
   - Are described muscle movements consistent with the depicted expression and basic facial anatomy?

3. Logical Coherence (1–5):
   - Are the reasoning steps internally consistent, non-contradictory, and clearly connected to the final FER label?
   - Does the explanation follow a clear, step-by-step structure from observations to conclusion?

You MUST:
- Judge only based on the given image, query, and the two responses.
- Ignore model identity and any stylistic preferences.
- Be strict about hallucinations and anatomically implausible claims.

--------------------------------
Input Format
--------------------------------
You will receive:

- The original user query describing the FER task.
- A face image `<image>`.
- Two anonymized model responses: **Response A** and **Response B**.

Each response may contain both reasoning and a final categorical prediction.

--------------------------------
Evaluation Instructions
--------------------------------
For each response, do the following:

1. Assign integer scores in [1, 5] for:
   - `visual_faithfulness_score`
   - `anatomical_precision_score`
   - `logical_coherence_score`

2. After scoring both responses, decide which one you overall prefer or whether they are tied:
   - `"A"` if Response A is clearly better overall,
   - `"B"` if Response B is clearly better overall,
   - `"tie"` if they are approximately equal.

Base your overall preference on the three rubric dimensions jointly, not just on the final label correctness.

--------------------------------
Output Format (Strict JSON Only)
--------------------------------
You MUST output ONLY a valid JSON object:

{
  "A": {
    "visual_faithfulness_score": 1-5,
    "anatomical_precision_score": 1-5,
    "logical_coherence_score": 1-5
  },
  "B": {
    "visual_faithfulness_score": 1-5,
    "anatomical_precision_score": 1-5,
    "logical_coherence_score": 1-5
  },
  "overall_preference": "A" | "B" | "tie",
  "justification": "Short paragraph explaining your scoring and preference, referencing concrete aspects of the two responses."
}
\end{Verbatim}
\end{promptbox}

%%%%%%%%%%%%%%%%%%%%%%%%%%%%%%%%%%%%%%%%%%%%%%%%%%%%%%%%%%%%

%\input{checklist}

\end{document}